\documentclass{article}%
\usepackage{amsmath}
\usepackage{amsfonts}
\usepackage{amssymb}
\usepackage{graphicx}
\usepackage{appendix}%
\begin{document}

\title{Predicting Regression Probability Distributions\\with Imperfect Data Through Optimal Transformations}
\author{Jerome H. Friedman\thanks{Department of Statistics, Stanford University, 390
Serra Mall, Stanford, CA 94305 (jhf@stanford.edu)}\\Stanford University\\\& Google, Inc.}
\maketitle

\begin{abstract}
The goal of regression analysis is to predict the value of a numeric outcome
variable $y$ given a vector of joint values of other (predictor) variables
$\mathbf{x}$. Usually a particular $\mathbf{x}$--vector does not specify a
repeatable value for $y$, but rather a probability distribution of possible
$y$--values, $p(y\,|\,\mathbf{x})$. This distribution has a location, scale
and shape, all of which can depend on\textbf{ }$\mathbf{x}$, and are needed to
infer likely values for $y$ given $\mathbf{x}$. Regression methods usually
assume that training data $y$-values are perfect numeric realizations from
some well behaived $p(y\,|\,\mathbf{x})$. Often actual training data
$y$-values are discrete, truncated and/or arbitrary censored. Regression
procedures based on an optimal transformation strategy are presented for
estimating location, scale and shape of $p(y\,|\,\mathbf{x})$ as general
functions of $\mathbf{x}$, in the possible presence of such imperfect training
data. In addition, validation diagnostics are presented to ascertain the
quality of the solutions.

Keywords: optimal transformations,\ heteroscedasticity, censoring, ordinal
regression, quantile regression

Running title: Predicting regression probability distributions

\end{abstract}

\section{Introduction}

In regression analysis one has a system under study with associated attributes
or variables. The goal is to estimate the unknown numeric (outcome) value of
one of the variables $y$, given the known joint values of other (predictor)
variables $\mathbf{x}=$ $(x_{1}\cdot\cdot\cdot,x_{p})$ associated with the
system. It is seldom the case that a particular set of $\mathbf{x}$--values
gives rise to a unique value for $y$. There are other variables $\mathbf{z}%
=(z_{1},z_{2},\cdot\cdot\cdot)$ that influence $y$ whose values are neither
controlled nor observed. Specifying a particular set of joint values for
$\mathbf{x}$ results in a probability distribution of possible $y$--values,
$p(y\,|\,\mathbf{x})$, induced by the varying values of $\mathbf{z}$. This
distribution has a location $f(\mathbf{x})$, scale $s(\mathbf{x})$ and shape,
all of which can depend on\textbf{ }$\mathbf{x}$. Using a training data set of
previously solved cases $\{y_{i},\mathbf{x}_{i}\}_{i=1}^{N}$ in which both the
outcome and predictor variable values are jointly observed, the goal is to
produce an estimate $\hat{p}(y\,|\,\mathbf{x})$ of the distribution of $y$
given $\mathbf{x}$.

Most regression procedures explicitly or implicitly approximate
$p(y\,|\,\mathbf{x})$ by a generic (usually normal) distribution that is
symmetric with constant scale, $s(\mathbf{x})=s$ (homoscedasticity). Only its
location function $f(\mathbf{x})$ is estimated from the training data. That
estimate $\hat{f}(\mathbf{x})$ can then be used to estimate the presumed
constant scale by%
\begin{equation}
\hat{s}=\sum_{i\in T}h(|\,y_{i}-\hat{f}(\mathbf{x}_{i})\,|) \label{e1}%
\end{equation}
on a \textquotedblleft test\textquotedblright\ data set $T$ not used to
estimate $\hat{f}(\mathbf{x})$. Here the particular function $h$ employed
depends upon the assumed distribution $p(y\,|\,\mathbf{x})$. In this setting a
$y$--prediction at a future $\mathbf{x}$--value is taken to be $\hat
{f}(\mathbf{x})$ since this is the most likely value of $y$ under a symmetric
distribution assumption. \ The uncertainty of all predictions is gauged by
$\hat{s}$ (\ref{e1}). The value of $\hat{s}$ is often taken to be a measure of
the lack-of-quality of the solution, especially in competitions.

The quantity $\hat{s}$ (\ref{e1}) reflects the prediction uncertainty averaged
over the marginal distribution of all $\mathbf{x}$--values, $p(\mathbf{x})$.
The actual uncertainty for predictions at a particular $\mathbf{x}$ is
characterized by $s(\mathbf{x})$, the scale of $p(y\,|\,\mathbf{x})$
corresponding to that $\mathbf{x}$--value. It is seldom the case that this
uncertainty is the same or even similar for different $\mathbf{x}$--values.
Generally there is substantial variation in the scale $s(\mathbf{x})$ over the
distribution of $\mathbf{x}$ (heteroscedasticity). When a predicted value of
$y$ is being used in an actual decision making application (rather than in a
competition) knowing its corresponding lack--of--accuracy can be important
information influencing the decision outcome.

It is important to note that the location $f(\mathbf{x})$ and scale
$s(\mathbf{x})$ are characteristics (parameters) of the distribution
$p(y\,|\,\mathbf{x})$. In particular, the scale function $s(\mathbf{x})$
represents an \textquotedblleft irreducible\textquotedblright\ error for
predictions at $\mathbf{x}$. Obtaining more training data or using more
effective learning algorithms cannot reduce this error. They can only reduce
the error in the estimated values of the parameters $\hat{f}(\mathbf{x})$ and
$\hat{s}(\mathbf{x})$. This latter \textquotedblleft
reducible\textquotedblright\ \ error is usually much smaller than the
irreducible error characterized by the intrinsic scale $s(\mathbf{x})$. The
only way to decrease irreducible error is to use a more informative set of
predictor variables $\mathbf{x}$.

Along with location and scale, higher moments can also vary with $\mathbf{x}$
giving rise to changing shape of $p(y\,|\,\mathbf{x})$ for different values of
$\mathbf{x}$. For example, $p(y\,|\,\mathbf{x})$ may be asymmetric with
varying degrees of skewness for different $\mathbf{x}$--values. This will
effect inference on likely values of $y$ for a given $\mathbf{x}$.

Almost all regression procedures implicitly assume that the outcome $y$ is a
real valued variable with the potential to realize any value for which the
marginal distribution of $y$%
\[
p(y)=\int p(y\,|\,\mathbf{x})\,\,p(\mathbf{x})\,d\mathbf{x}%
\]
has support. For most hypothesized $p(y\,|\,\mathbf{x})$ used in regression
analysis this implies all real values $y\in R$. In many applications, however,
this ideal is not realized. Recorded $y$--values may be restricted to a small
distinct set with many ties. The tied values themselves may be unknown. Only
an order relation among them is recorded (ordinal regression). In other
applications, one may only be able to specify (possibly overlapping) intervals
that contain each training data $y$--value, rather than the actual value
itself (censoring). The specified intervals may be open or closed.

This paper describes an omnibus regression procedure (OmniReg) that can use
imperfect training data such as described above to produce estimates of the
location $f(\mathbf{x})$, scale $s(\mathbf{x})$ and shape of
$p(y\,|\,\mathbf{x})$ as general functions of $\mathbf{x}$. These can be used
to assess the distribution of likely values of $y$ for new observations for
which only the predictor values $\mathbf{x}$ are known.

Section \ref{est} presents the basic gradient boosting strategy for jointly
estimating the location $f(\mathbf{x})$ and scale $s(\mathbf{x})$ of a
symmetric $p(y\,|\,\mathbf{x})\,$\ in the presence of general censoring of the
outcome variable $y$. Section \ref{opt} generalizes the procedure by
presenting a method for for constructing the optimal transformation of $y$,
$g(y)$, on which to perform the corresponding location--scale estimation.
Section \ref{diag} outlines diagnostic procedures for checking the consistency
of derived solutions. Section \ref{asym} further generalizes the procedure by
incorporating asymmetry, as a function of $\mathbf{x}$, into the transformed
$p(g(y)\,|\,\mathbf{x})$ solutions. Illustrations using three publicly
available data sets, one censored and two uncensored, are presented in Section
\ref{examp}. Application to ordinal regression, where only an order relation
among the $y$-values is observed, is discussed in Section \ref{ord}.
Connections of this method to other related techniques are discussed in
Section \ref{rel}. Section \ref{point} outlines the use of the derived
distribution estimates $\hat{p}(y\,|\,\mathbf{x})$ for point prediction.
Section \ref{disc} provides a concluding discussion.

\section{Estimation\label{est}}

Following Tobin (1958) we assume that the outcome $y$ is a possibly imperfect
measurement of a random variable $y^{\ast}$ that follows a well behaved
probability distribution $p(y^{\ast}\,|\,\mathbf{x})$ with support on the
entire real line. In particular, we initially suppose that $y^{\ast}$ for a
given $\mathbf{x}$ follows an additive error model%
\begin{equation}
y^{\ast}=\,f(\,\mathbf{x})+s(\mathbf{x})\cdot\varepsilon\label{e1.5}%
\end{equation}
with location $f(\mathbf{x})$ and scale $s(\mathbf{x)}$, where both are
general functions of $\,\mathbf{x}$ to be estimated from the training data.
The random variable $\varepsilon$ follows a standard logistic distribution
\begin{equation}
\varepsilon\sim\frac{\exp(-\varepsilon\mathbf{)}}{(1+\exp(-\varepsilon))^{2}%
}\text{.} \label{e2}%
\end{equation}
This distribution has a Gaussian shape near its center and continuously
evolves to that of a Laplace distribution in the tails. This provides
robustness to potential outliers while maintaining power for nearly normally
distributed data.

\subsection{Data\label{data}}

The data imperfections described above can all be treated as being different
aspects of censoring. The outcome value for a censored observation is unknown;
one can only specify an interval containing its corresponding value $y^{\ast
}\in\lbrack a,b]$. If $a=b$ the observation is uncensored. If $a=-\infty$ the
observation is said to be left censored at $b$; if $b=\infty$ it is right
censored at $a$. Otherwise it is interval censored. An actual data set can
consist of a mixture of all of these censored types with or without the
inclusion of uncensored observations. In the case of discrete $y$--values with
ties, one can consider the observations at each tied value to be interval
censored between the midpoint of the previous and next set of possibly tied values.

\subsection{Loss function\label{loss}}

We use maximum likelihood based on the logistic distribution (\ref{e1.5})
(\ref{e2}) to estimate the location $f(\mathbf{x})$ and scale $s(\mathbf{x)}$
functions from the training data $\{y_{i},\mathbf{x}_{i}\}_{i=1}^{N}$ . The
$y$--value for each observation $i$ is specified by a lower bound $a_{i}$ and
upper bound $b_{i}$. If $a_{i}<b_{i}$ the probability that $y_{i}^{\ast}%
\in\lbrack a_{i},b_{i}]$ is given by%
\[
\Pr(y_{i}^{\ast}\in\lbrack a_{i},b_{i}])=\frac{1}{1+\exp((f(\mathbf{x}%
_{i})-b_{i})/s(\mathbf{x}_{i}))}-\frac{1}{1+\exp((f(\mathbf{x}_{i}%
)-a_{i})/s(\mathbf{x}_{i}))}\text{.}%
\]
If $a_{i}=b_{i}$, it is given by%
\begin{equation}
\Pr(y_{i}^{\ast}=b_{i})=\frac{1}{s(\mathbf{x}_{i}\mathbf{)}}\frac
{\exp((f(\mathbf{x}_{i})-b_{i})/s(\mathbf{x}_{i}))}{(1+\exp((f(\mathbf{x}%
_{i})-b_{i})/s(\mathbf{x}_{i})))^{2}}\text{.} \label{e2.5}%
\end{equation}
The loss function then becomes the negative log--likelihood%
\begin{align}
L(a,b,f(\mathbf{x}),s(\mathbf{x}))  &  =I(a=b)\,[\,\log(s(\mathbf{x}%
))+(b-f(\mathbf{x}))/s(\mathbf{x})+2\,\log(1+\exp((f(\mathbf{x}%
)-b)/s(\mathbf{x}))]\nonumber\\
-I(a  &  <b)\,\log\,\left[  \frac{1}{1+\exp((f(\mathbf{x})-b)/s(\mathbf{x}%
))}-\frac{1}{1+\exp((f(\mathbf{x})-a)/s(\mathbf{x}))}\right]  \text{.}
\label{e3}%
\end{align}
The function estimates are then%
\begin{equation}
(\hat{f}(\mathbf{x}),\hat{s}(\mathbf{x}))=\arg\min_{f,s\in F}\sum_{i=1}%
^{N}L(a_{i},b_{i},f(\mathbf{x}_{i}),s(\mathbf{x}_{i})) \label{e4}%
\end{equation}
where $F$ is some chosen class of functions.

\subsection{Implementation\label{impl}}

The function class $F$ in (\ref{e4}) employed here consists of linear
combinations of decision trees induced by gradient boosting (Friedman 2001)%
\begin{equation}
\hat{f}(\mathbf{x})=\sum_{k=1}^{K_{f}}T_{k}^{(f)}(\mathbf{x}) \label{e5}%
\end{equation}
and%
\begin{equation}
\widehat{\log(s(\mathbf{x}))}=\sum_{k=1}^{K_{s}}T_{k}^{(s)}(\mathbf{x}).
\label{e6}%
\end{equation}
Each $T_{k}^{(f)}(\mathbf{x})$ and $T_{k}^{(s)}(\mathbf{x})$ is a different
CART$^{TM}$ decision tree (Breiman \emph{et. al. }1984) using $\mathbf{x}$ as
predictors. The trees in each expansion (\ref{e5}) (\ref{e6}) are sequentially
induced using the respective generalized residuals%
\begin{equation}
\tilde{r}_{f}(a,b,f\,|\,s)=-\frac{\partial L(a,b,f,s)}{\partial f} \label{e7}%
\end{equation}
and%
\begin{equation}
\tilde{r}_{s}(a,b,s\,|\,f)=-\frac{\partial L(a,b,f,s)}{\partial\log(s)}
\label{e8}%
\end{equation}
as the outcome with the loss function $L(a,b,f,s)$ is given by (\ref{e3}). At
each step the evaluation of $f(\mathbf{x})$ and $s(\mathbf{x})$ is based on
the previously induced trees in the respective sequences. The quantity
$\log(s(\mathbf{x}))$ is estimated in (\ref{e6}) because the loss function
(\ref{e3}) is not convex in $s$, but is convex in $\log(s)$. This also
enforces the constraint that the estimated scale $\hat{s}(\mathbf{x}%
)=\exp(\widehat{\log(s(\mathbf{x}))})$ is always greater than zero.

After each tree in (\ref{e5}) is built the predicted value in each of its
terminal nodes $t$ is taken to be $\eta\cdot\tilde{f}_{t}$ where $\tilde
{f}_{t}$ is the solution to the line search
\begin{equation}
\sum_{i\in t}\tilde{r}_{f}(a_{i},b_{i},\hat{f}\,(\mathbf{x}_{i})+\tilde{f}%
_{t}\,|\,\hat{s}(\mathbf{x}_{i}))=0\text{,} \label{e9}%
\end{equation}
and $0<\eta<<1$ is a small number (learning rate). For the trees in (\ref{e6})
the terminal node predictions are given by $\eta\cdot\log(\tilde{s}_{t})$
with
\begin{equation}
\sum_{i\in t}\tilde{r}_{s}(a_{i},b_{i},\hat{s}(\mathbf{x}_{i})\,\cdot\tilde
{s}_{t}\,|\,\hat{f}(\mathbf{x}_{i}))=0. \label{e10}%
\end{equation}
In (\ref{e9}) (\ref{e10}) $\hat{f}\,(\mathbf{x})$ and $\hat{s}(\mathbf{x})$are
the current location and scale function estimates used to induce the
corresponding tree.

The number of trees $K_{f}$ and $K_{s}$ in (\ref{e5}) and (\ref{e6})
respectively are taken to be those that minimize the negative log--likelihood
(\ref{e4}) as evaluated on an independent \textquotedblleft
test\textquotedblright\ set of observations not used to learn $\hat
{f}\,(\mathbf{x})$ and $\hat{s}(\mathbf{x})$.

The tree sequences (\ref{e5}) (\ref{e6}) are alternatively induced using an
iterative boosting algorithm:\newline

\qquad\qquad\qquad\qquad\qquad ITERATIVE GRADIENT BOOSTING

\qquad$\qquad\qquad\qquad\qquad$Start: $\hat{s}(\mathbf{x})=$ constant

\qquad\qquad\qquad\qquad\ \qquad Loop \{

\qquad\qquad$\qquad\qquad\qquad\qquad\hat{f}(\mathbf{x})=\,$tree--boost
$f(\mathbf{x})$ given $\hat{s}(\mathbf{x})$

\qquad\qquad$\qquad\qquad\qquad\qquad\widehat{\log(s(\mathbf{x}))}%
=\,$tree--boost $s(\mathbf{x})$ given $\hat{f}(\mathbf{x})$

\qquad\qquad\qquad\qquad\ \qquad\}\thinspace Until change
$<$
threshold.\newline\newline

Starting with $\hat{s}(\mathbf{x})$ being a constant function, gradient
boosting is applied to estimate a corresponding $\hat{f}(\mathbf{x})$ using
(\ref{e7}) (\ref{e9}). Given that $\hat{f}(\mathbf{x})$ boosting is applied to
estimate its corresponding $\log(\hat{s}(\mathbf{x}))$ using (\ref{e8})
(\ref{e10}). This $\hat{s}(\mathbf{x})$ is then used to estimate a new
$\hat{f}(\mathbf{x})$ and so on. The alternating iterative process continues
until the change in both functions on successive iterations is below a small
threshold. Usually only two or three iterations are required.

\section{Optimal transformations\label{opt}}

The strategy described in Section \ref{est} attempts to capture location
$f(\mathbf{x})$ and scale $s(\mathbf{x})$ as functions of the predictor
variables $\mathbf{x}$ under an additive error assumption (\ref{e1.5}).
However, there is no guarantee that the actual $p(y\,|\,\mathbf{x})$ in any
application even approximately has this property. Violation can result in
highly distorted estimates $\hat{p}(y\,|\,\mathbf{x})$. While accurate
probability estimates are essential for proper inference, they are especially
important for estimation in the presence of censoring (Section \ref{data}).
Given a censoring interval $[a,b]$, knowledge concerning the corresponding
underlying outcome value $y^{\ast}$ is derived solely from its estimated distribution.

One way to mitigate this problem is by transforming the outcome variable $y$
using a monotonic function. The goal here is to find a monotonic
transformation function $g(y)$ such that (\ref{e1.5}) at least approximately
holds for the transformed variable. That is%
\begin{equation}
g(y)\simeq\,f(\,\mathbf{x})+s(\mathbf{x})\cdot\varepsilon. \label{e10.5}%
\end{equation}
To the extent (\ref{e10.5}) holds, $g(y)$ represents the best transformation
for estimating the corresponding $f(\,\mathbf{x})$ and $s(\mathbf{x})$. All
inference can be performed on the transformed variable $z=g(y)$ using the
distribution of $\varepsilon$ (\ref{e2}). Corresponding $p$-quantiles
referencing the distribution of the original untransformed variable $y$,
$p(y\,|\,\mathbf{x})$, $\ $are given by $q_{p}(y\,|\,\mathbf{x})=g^{-1}%
[q_{p}(z\,|\,\mathbf{x})]$.

For any given transformation $g(y)$ one can directly solve (\ref{e4}) for its
corresponding $\hat{f}(\mathbf{x})$ and $\hat{s}(\mathbf{x})$, using the
methods described in Section \ref{impl}, by simply taking $g(y)$ to be the
outcome variable in place of $y$. That is $a_{i}\rightarrow g(a_{i})$ and
$b_{i}\rightarrow g(b_{i})$. Since $g(y)$ is monotonic its cumulative
distribution $G(g(y))$ for any $y$ must be the same as that of its
corresponding untransformed outcome $y$, $F(y)$. That is,
$G(g(y)\,|\,\mathbf{x})=$ $F(y\,|\,\mathbf{x})$ at any $\mathbf{x}$ so that
their respective data averages are equal%
\begin{equation}
\frac{1}{N}\sum_{i=1}^{N}G(g(y)\,|\,\mathbf{x}_{i}))=\frac{1}{N}\sum_{i=1}%
^{N}F(y\,|\,\mathbf{x}_{i}))\text{.} \label{e10.6}%
\end{equation}
Given $F$ and $G$ this (\ref{e10.6}) defines the optimal transformation
$g(y)\,$.

The right hand side of (\ref{e10.6}) can be estimated by the empirical
marginal cumulative distribution of $y$, $\hat{F}(y)$. By assumption
(\ref{e10.5}) the cumulative distribution of $g(y)$ at each $\mathbf{x}$ is%
\begin{equation}
G(g(y)\,|\,\mathbf{x})=\frac{1}{1+\exp((f(\mathbf{x})-g(y))/s(\mathbf{x}%
))}\text{.} \label{e10.7}%
\end{equation}
Substituting the corresponding estimates for the distribution parameters one
has from (\ref{e10.6})%

\begin{equation}
\frac{1}{N}\sum_{i=1}^{N}\frac{1}{1+\exp((\hat{f}(\mathbf{x})-\hat{g}%
(y))/\hat{s}(\mathbf{x}))}=\hat{F}(y)\text{.} \label{e10.9}%
\end{equation}
Given any value for $y$ one can solve (\ref{e10.9}) for its corresponding
estimated transformed value $\hat{g}(y)$ using a line search method. Note that
$\hat{g}(y)$ as defined by (\ref{e10.9}) is only required to be monotonic and
is not restricted to any other function class.

If all of the training data $y$-values are uncensored, or the censoring
intervals are restricted to non--overlapping bins, then the corresponding
$\hat{F}(y)$ is trivially obtained by ranking the training data $y$-values. If
not, it can be estimated using Turnbull's non--parametric self--consistency
algorithm (Turnbull 1976).

These considerations suggest an alternating optimization algorithm to jointly
solve for the functions $\hat{g}(y)$, $\hat{f}(\mathbf{x})$ and $\hat
{s}(\mathbf{x})$. Starting with an initial guess for $\hat{g}(y)$ (say
$\hat{g}(y)=y$), one solves (\ref{e3}) (\ref{e4}) for the corresponding
$\hat{f}(\mathbf{x})$ and $\hat{s}(\mathbf{x})$ as described in Section
\ref{impl} using the current $\hat{g}(y)$ as the outcome variable. Given the
resulting solution location and scale functions one evaluates a new $\hat
{g}(y)$ using (\ref{e10.9}). This transformation function replaces the
previous one resulting in new estimates for $\hat{f}(\mathbf{x})$ and $\hat
{s}(\mathbf{x})$ from (\ref{e3}) (\ref{e4}). These in turn produce a new
transformation through (\ref{e10.9}) and so on. The process is continued until
the estimates stop changing, converging to a fixed point. Usually three to
five iterations are sufficient.

\section{Diagnostics\label{diag}}

No procedure works equally well in all applications. Simply because a program
returns a result does not insure its validity. It is important to assess
whether or not the results accurately reflect their corresponding population
quantities by subjecting them to diagnostic consistency checks or constraints.
To the extent these constraints are satisfied they provide necessary, but not
sufficient, evidence for the validity of the model. Those that are violated
uncover corresponding model inadequacies. In this section diagnostics are
presented for checking the consistency of the probability distribution
$\hat{p}(y\,|\,\mathbf{x})$ estimates.

\subsection{Marginal distribution plots\label{plots}}

The fundamental premise of the procedure is that given any $\mathbf{x}$, its
corresponding transformed outcome $\hat{g}(y)$ is a random variable that
follows a symmetric logistic distribution with location $\hat{f}(\mathbf{x})$
and scale $\hat{s}(\mathbf{x})$. If so, its corresponding cumulative
distribution is%
\begin{equation}
\Pr(\hat{g}(y)<z)=1/(1+\exp((\hat{f}(\mathbf{x})-z)/\hat{s}(\mathbf{x}%
)))\text{.} \label{e19}%
\end{equation}
It is not possible to verify (\ref{e19}) for any single observation
$\mathbf{x}=\mathbf{x}_{i}$. However it can be tested for subsets of data. Let
$S$ be a subset of observations drawn from a validation data set not used to
obtain the estimates $\hat{g}(y)$, $\hat{f}(\mathbf{x})$ or $\hat
{s}(\mathbf{x})$. The subset $S$ must be selected using only predictor
variable values $\mathbf{x}$, or quantities derived from them, and not involve
the outcome $y$. Then the predicted cumulative distribution of $\hat{g}(y)$
for $\mathbf{x}\in S$ is%
\begin{equation}
\Pr(\hat{g}(y)<z)=\frac{1}{N_{S}}\sum_{\mathbf{x}_{i}\in S}1/(1+\exp((\hat
{f}(\mathbf{x}_{i})-z)/\hat{s}(\mathbf{x}_{i})))\text{.} \label{e20}%
\end{equation}
One can compare this to the actual empirical distribution of $\hat{g}(y)$ for
$\mathbf{x}\in S$. This \ can be computed by sorting the $\{y_{i}%
\}_{\mathbf{x}_{i}\in S}$ or by using Turnbull's non--parametric algorithm if
there is censoring present. The size of the subset $N_{S}$ should be large
enough to obtain reliable estimates of the empirical distribution.

The empirical and predicted distributions can be compared with
quantile--quantile (Q-Q) plots. The vertical axis of a Q-Q plot represents the
empirical quantiles of the data subset. The abscissa represents the same
quantiles based on the predicted distribution (\ref{e20}). To the extent that
the two distributions are the same, points corresponding to the same quantile
value will tend to be equal and thus lie close to a 45--degree straight line.
If the two distributions have the same shape and scale but differ in location
then the corresponding quantiles will lie close to a line parallel to the
diagonal but shifted by the location difference. If the shape of the two
distributions is the same but they differ in scale the corresponding quantiles
will lie on a straight line with a non unit slope reflecting the differing
scales. To the extent the shapes of the two distributions differ the
respective quantiles will fail to lie on a straight line.

If the OmniReg model is correct the prediction (\ref{e20}) holds for any
$\mathbf{x}$-defined data subset $S$. It is not feasible to test it over all
possible subsets of the data to look for discrepancies. One must judiciously
choose revealing subsets, perhaps based on domain knowledge. One generic
possibility is to select subsets based on joint values of the estimated
location $\hat{f}(\mathbf{x})$ and scale $\hat{s}(\mathbf{x})$. That is
$S=\{i\,|\,r<\hat{f}(\mathbf{x}_{i})\leq t\;\&\;\,u<\hat{s}(\mathbf{x}%
_{i})\leq v\}$. This approach is illustrated on the data example in Section
\ref{quest}.

\subsection{Standardized residual plots\label{stdres}}

The marginal distribution diagnostics described in Section \ref{plots} can be
computed for data containing mixtures of uncensored and any kind of censored
outcomes. However if all outcomes in the validation data set are uncensored,
the predicted standardized residuals%
\begin{equation}
\hat{r}(y\,|\,\mathbf{x})=(\hat{g}(y)-\hat{f}(\mathbf{x}))/\hat{s}(\mathbf{x})
\label{e29}%
\end{equation}
can be directly computed and examined. It is these standardized residuals that
are central to inference. Under procedure assumptions, and to the extent that
the transformation $\hat{g}(y)$, location $\hat{f}(\mathbf{x})$ and scale
$\hat{s}(\mathbf{x}_{i})$ function estimates are accurate, the predicted
standardized residuals (\ref{e29}) follow a standard logistic distribution
(\ref{e2}) for any $\mathbf{x}$. As above this can't be verified for any
single observation $\mathbf{x}=\mathbf{x}_{i}$, but it can be tested for
groups of data $\mathbf{x}\in S$ as described in Section \ref{plots}. These
diagnostics are illustrated on the data examples in Sections \ref{mash} and
\ref{mds}.

\section{Asymmetry\label{asym}}

The optimal transformation strategy (Section \ref{opt}) attempts to find a
monotonic function $g(y)$ such that (\ref{e10.5}) approximately holds. To the
extent it is successful the transformed distribution $p(g(y)\,|\,\mathbf{x})$
must be symmetric or close to it. Although it often succeeds, there is no
guarantee that such a transformation exists in a particular application. When
that happens the procedure can compromise accuracy of function estimates by
attempting to approximate distribution symmetry. This will likely be reflected
in the diagnostic plots described in Sections \ref{plots} and \ref{stdres}.

\begin{figure}
[ptb]
\begin{center}
\includegraphics[
height=3.269in,
width=4.894in
]%
{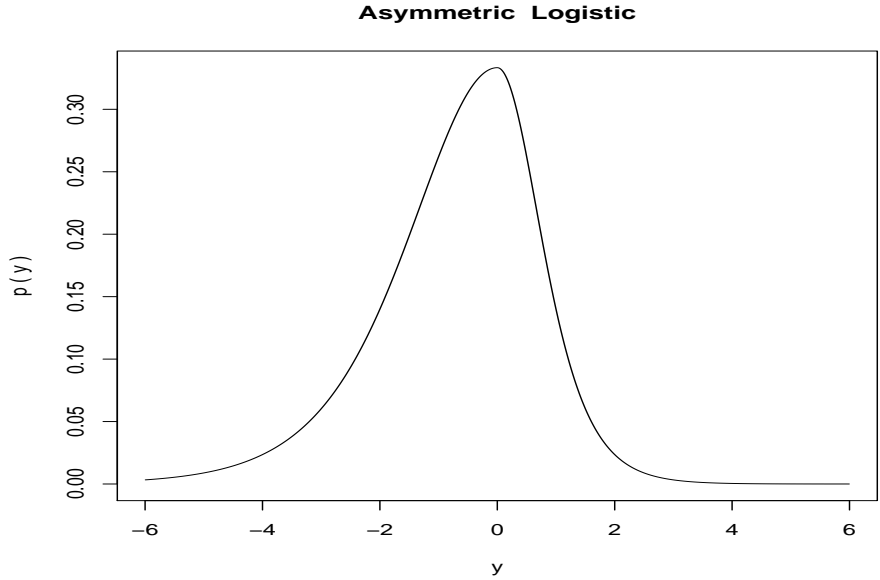}%
\caption{Probability density of asymmetric logistic distribution (20) with
mode $f$ $=0$, lower scale $s_{l}=2$ and upper scale $s_{u}=1$.}%
\label{fig20}%
\end{center}
\end{figure}

A way to mitigate this problem is to relax the symmetry requirement on
$p(g(y)\,|\,\mathbf{x})$ in the transformed setting. In particular the
logistic distribution can be generalized to incorporate asymmetry by providing
different scales on the positive and negative residuals%
\begin{equation}
p(z\,|\,f,s_{l},s_{u})=\frac{2\,}{s_{l}+s_{u}}\left[  \frac{I(z\leq
f)\,\exp((f-z)/s_{l})}{(1+\exp((f-z)/s_{l}))^{2}}+\frac{I(z>f)\,\exp
((f-z)/s_{u})}{(1+\exp((f-z)/s_{u}))^{2}}\right]  \text{.} \label{e30}%
\end{equation}
Here $f$ represents the mode, $s_{l}$ a scale parameter for the negative
residuals and $s_{u}$ a corresponding scale for the positive residuals. Figure
\ref{fig20} shows a plot of (\ref{e30}) for $f=0$, $s_{l}=2$, $s_{u}=1$.%

Note that this definition of an asymmetric logistic distribution is different
from the \textquotedblleft skew\textquotedblright\ logistic distributions
proposed by Johnson \emph{et. al.} (1995). It (\ref{e30}) is faster to compute
in that it only requires the evaluation of a single exponential function,
whereas the skew logistic requires two exponentials and a logarithm. Also, its
parameters have a straightforward interpretation.

The iterative gradient boosting strategy of Section \ref{impl} is easily
modified to incorporate this generalization. The probability density
(\ref{e30}) and its cumulative distribution%
\begin{equation}
CDF(z\,|\,f,s_{l},s_{u})=\frac{2\,s_{l}}{s_{l}+s_{u}}\left\{  \frac{I(z\leq
f)}{1+\exp((f-z)/s_{l})}+\,I(z>f)\left[  \frac{1}{2}+\frac{s_{u}}{s_{l}%
}\left(  \frac{1}{1+\exp((f-z)/s_{u}))}-\frac{1}{2}\right)  \right]  \right\}
\label{e31}%
\end{equation}
simply replace the corresponding symmetric logistic distributions in the
formulation of a loss function $L(z\,|\,f,s_{l},s_{u})$ analogous to
(\ref{e3}). Three functions $f(\mathbf{x})$, $s_{l}(\mathbf{x})$ and
$s_{u}(\mathbf{x})$ are estimated by gradient boosting:

\qquad\qquad\qquad\qquad\quad ASYMMETRIC GRADIENT BOOSTING

\qquad\qquad\qquad\qquad Start: $\hat{s}_{l}(\mathbf{x})=\hat{s}%
_{u}(\mathbf{x})=$ constant

\qquad\qquad\qquad\qquad Loop \{

\qquad\qquad\qquad\qquad$\quad\hat{f}(\mathbf{x})=\,$tree--boost
$f(\mathbf{x})$ given $\hat{s}_{l}(\mathbf{x})\,\&\,\hat{s}_{u}(\mathbf{x})$

$\qquad\qquad\qquad\qquad\quad\widehat{\log(s_{l}(\mathbf{x}))}=\,$tree--boost
$s_{l}(\mathbf{x})$ given $\hat{f}(\mathbf{x})\,\&\,\hat{s}_{u}(\mathbf{x})$

$\qquad\qquad\qquad\qquad\widehat{\quad\log(s_{u}(\mathbf{x}))}=\,$tree--boost
$s_{u}(\mathbf{x})$ given $\hat{f}(\mathbf{x})\,\&\,\hat{s}_{l}(\mathbf{x})$

\qquad\qquad\qquad\quad\quad\}\thinspace Until change
$<$
threshold\newline

\qquad\newline For the optimal transformation strategy of Section \ref{opt},
(\ref{e31}) simply replaces (\ref{e10.7}) in (\ref{e10.9}).

A limitation of this asymmetric gradient boosting algorithm is that its
convergence is quite slow. This is caused by the strong effect that each scale
parameter has on the solution for the other scale parameter. However in the
special case of (\ref{e30}) with uncensored outcomes there is a trick that
dramatically speeds convergence. This approach is based on the symmetric
logistic loss function (\ref{e3}) with $a=b=y$.

At each $\mathbf{x}$ there are estimates for the three functions $\hat
{f}(\mathbf{x})$, $\hat{s}_{l}(\mathbf{x})$ and $\hat{s}_{u}(\mathbf{x})$.
Associated with each training observation $i$ is a location $f_{i}=$ $\hat
{f}(\mathbf{x}_{i})$\ and \emph{single} scale parameter $s_{i}$. If $y_{i}%
\leq$ $\hat{f}(\mathbf{x}_{i})$ then $s_{i}=\hat{s}_{l}(\mathbf{x}_{i})$,
otherwise $s_{i}=\hat{s}_{u}(\mathbf{x}_{i})$. At each iteration of asymmetric
boosting a new location function $\hat{f}(\mathbf{x})$ is estimated based on
all of the training data $\{y_{i},f_{i},s_{i}\}_{i=1}^{N}$. Next a new lower
scale function $s_{l}(\mathbf{x})$ is estimated using only the training data
with currently negative residuals $\{y_{i},f_{i},s_{i}\}_{y_{i}\leq f_{i}}$.
Then a new upper scale function $\hat{s}_{u}(\mathbf{x})$ is estimated using
only the training data with currently positive residuals $\{y_{i},f_{i}%
,s_{i}\}_{y_{i}>f_{i}}$. All estimation is based on the symmetric logistic
loss function (\ref{e3}). This leads to an alternative algorithm for
uncensored $y$:

\qquad

\qquad\qquad\qquad\qquad\quad ASYMMETRIC GRADIENT BOOSTING (2)

\qquad\qquad\qquad\qquad Start: $\hat{s}_{l}(\mathbf{x})=\hat{s}%
_{u}(\mathbf{x})=$ constant

\qquad\qquad\qquad\qquad Loop \{

\qquad\qquad\qquad\qquad$\quad\hat{f}(\mathbf{x})=\,$tree--boost
$f(\mathbf{x})$ given $\hat{s}_{l}(\mathbf{x})\,\&\,\hat{s}_{u}(\mathbf{x})$

$\qquad\qquad\qquad\qquad\quad\widehat{\log(s_{l}(\mathbf{x}))}=\,$tree--boost
$s_{l}(\mathbf{x})$ given $\,y\leq\hat{f}(\mathbf{x})\,$

$\qquad\qquad\qquad\qquad\widehat{\quad\log(s_{u}(\mathbf{x}))}=\,$tree--boost
$s_{u}(\mathbf{x})$ given $y>\hat{f}(\mathbf{x})\,$

\qquad\qquad\qquad\quad\quad\}\thinspace Until change
$<$
threshold.\newline

This algorithm uncouples the direct effect of each scale function \ on the
estimation of the other. They only interact indirectly through their effect on
the location function estimate. Convergence usually requires four to six iterations.

The diagnostic plots of Section \ref{plots} are easily generalized to
asymmetric error model. The asymmetric cumulative distribution (\ref{e31})
simply replaces (\ref{e19}). For the standardized residual plots (Section
\ref{stdres}) an asymmetrically standardized residual%
\begin{equation}
\hat{r}_{a}(y\,|\,\mathbf{x})=I\,[\hat{g}(y)\leq\hat{f}(\mathbf{x}%
)]\,\,(\hat{g}(y)-\hat{f}(\mathbf{x})\,)/\hat{s}_{l}(\mathbf{x})+I\,[\hat
{g}(y)>\hat{f}(\mathbf{x})]\,\,(\hat{g}(y)-\hat{f}(\mathbf{x})\,)\,/\hat
{s}_{u}(\mathbf{x}) \label{e35}%
\end{equation}
replaces the symmetric one (\ref{e29}).

\section{Data examples\label{examp}}

In this section the procedures described in the previous sections are
illustrated using three data sets. The first involves censoring in which none
of the actual outcome $y$-values are known. The other two are well known
regression data sets from the Irvine Machine Learning Repository and involve
uncensored outcomes. For all data sets two analyses are performed. In the
first, location $\hat{f}(\mathbf{x})$ and scale $\hat{s}(\mathbf{x})$
estimates are derived assuming that the underlying variable $y^{\ast}$ follows
a symmetric logistic distribution (\ref{e1.5}) (\ref{e2}) as described in
Section \ref{est}, without the application of any transformation. In the
second analysis the procedure described in Section \ref{opt} is applied to
jointly estimate the optimal transformation $g(y)$ along with its
corresponding location and scale functions. In addition, an asymmetric
analysis (Section \ref{asym}) is applied to the third data set. All presented
results are based on predictions using a validation data set not involved in
model construction or selection.

\subsection{Questionnaire data\label{quest}}

This data set contains demographics on people who filled out questionnaires at
shopping malls in the San Francisco Bay Area (Hastie, Tibshirani, and Friedman
2009). The exercise here is to predict a person's age given the other
demographic information listed on their questionnaire. The predictor variables
are listed in Table 2.

\begin{center}
\textbf{Table 2}

Questionnaire predictor variables

\ 

$%
\begin{tabular}
[c]{ll}%
1 & Occupation\\
2 & Type of home\\
3 & Gender\\
4 & Martial status\\
5 & Education\\
6 & Annual income\\
7 & Lived in Bay Area\\
8 & Dual Incomes\\
9 & Persons in household\\
10 & Persons in household under 18\\
11 & Householder status\\
12 & Ethnic classification\\
13 & Language
\end{tabular}
$
\end{center}

Questionnaire age values are specified as being in one of seven intervals as
shown in Table \ 3.

\begin{center}
\textbf{Table 3}

Specified age intervals

\ %

\begin{tabular}
[c]{ccccccc}%
1 & 2 & 3 & 4 & 5 & 6 & 7\\
17 and under & 18--24 & 25--34 & 35--44 & 45--54 & 55--64 & 65 and older
\end{tabular}

\end{center}

This outcome can be considered as censored in non overlapping intervals with
bounds $\ B=\{0,17.5,24.5,34.5,44.5,54.5,64.5,\infty\}$. There are $N=8856$
questionnaires. These are randomly divided into $5000$ observations for model
construction, $2000$ for selecting the number of trees in each model
\ (\ref{e5}) (\ref{e6}) , and $1856$ for validation.%

\begin{figure}
[ptb]
\begin{center}
\includegraphics[
height=8.1707in,
width=5.994in
]%
{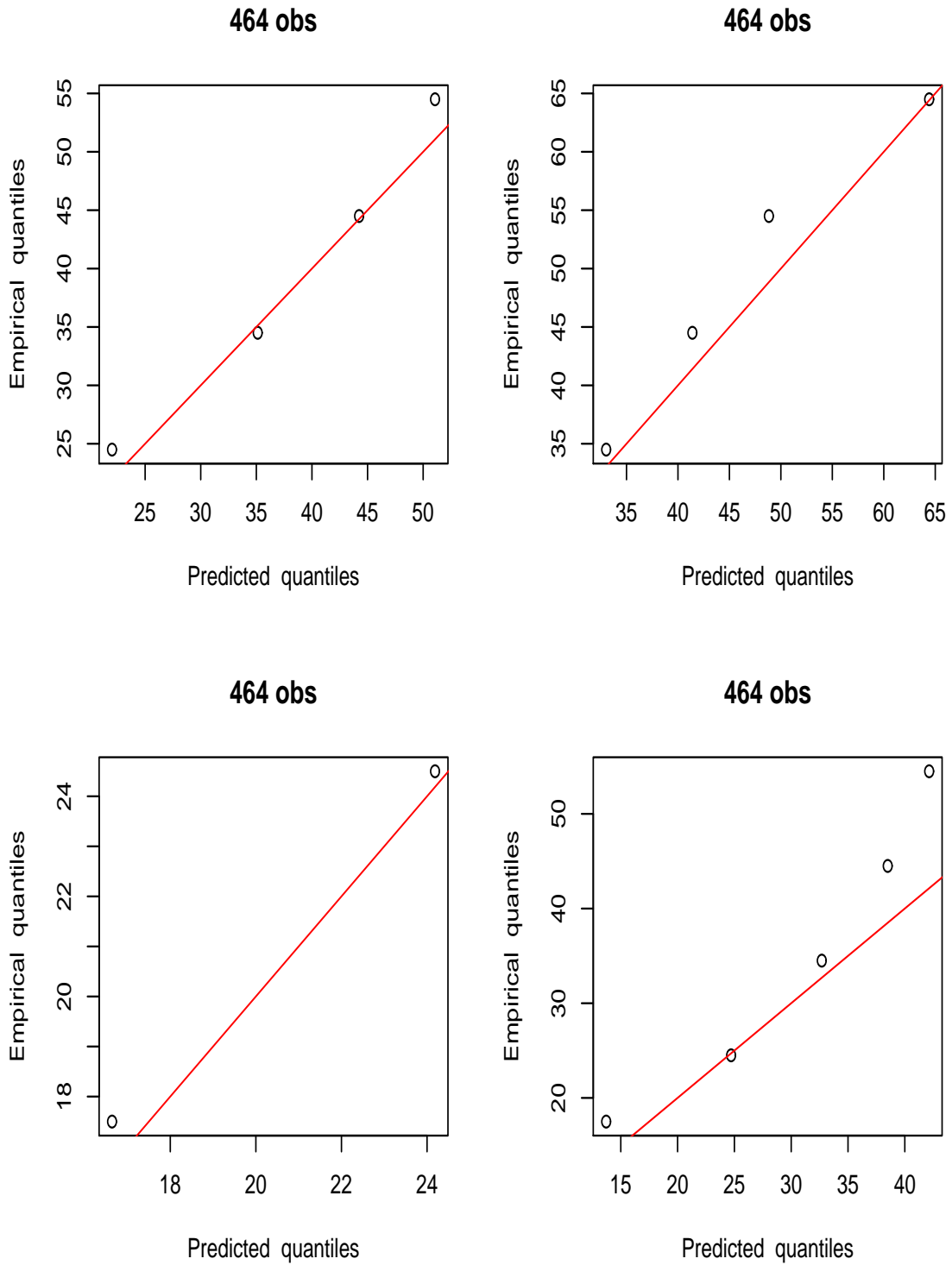}%
\caption{Questionnaire data. Q--Q plots of empirical versus predicted
quantiles for untransformed Omnireg solution for four data subsets obtained by
partitioning the location $\hat{f}(\mathbf{x})$ and $\hat{s}(\mathbf{x})$
estimates at their respective medians.}%
\label{fig5}%
\end{center}
\end{figure}

Figure \ref{fig5} shows four Q--Q plots of the empirical versus predicted
quantiles of the marginal distribution of age (Section \ref{plots}) based on
the untransformed solution. Each plot represents data within a joint interval
of $\hat{f}(\mathbf{x}_{i})$ and $\hat{s}(\mathbf{x}_{i})$ values. The
interval boundaries for $\hat{f}(\mathbf{x}_{i})$ are obtained by partitioning
at its corresponding median (rows -- bottom to top). Within each such location
interval separate intervals for $\hat{s}(\mathbf{x}_{i})$ are constructed
using its median evaluated within that location interval (columns -- left to
right). Thus the lower left plot is based on the \ 25\% of the observations
with the smallest locations and smallest scales. The top right involves the
25\% of the observations with jointly the largest estimated locations and
scales. Because of the censoring this marginal distribution is observable only
at the six values that separate the censoring intervals. In Fig. \ref{fig5},
points representing extreme empirical \ quantiles based on less than $20$
observations are not shown due to their instability. One sees In Fig.
\ref{fig5} that the predicted quantiles approximately follow the actual
empirical quantiles. However there are a few clear discrepancies in the right
plots representing the larger scale estimates.

\begin{figure}
[ptb]
\begin{center}
\includegraphics[
height=3.6313in,
width=4.9208in
]%
{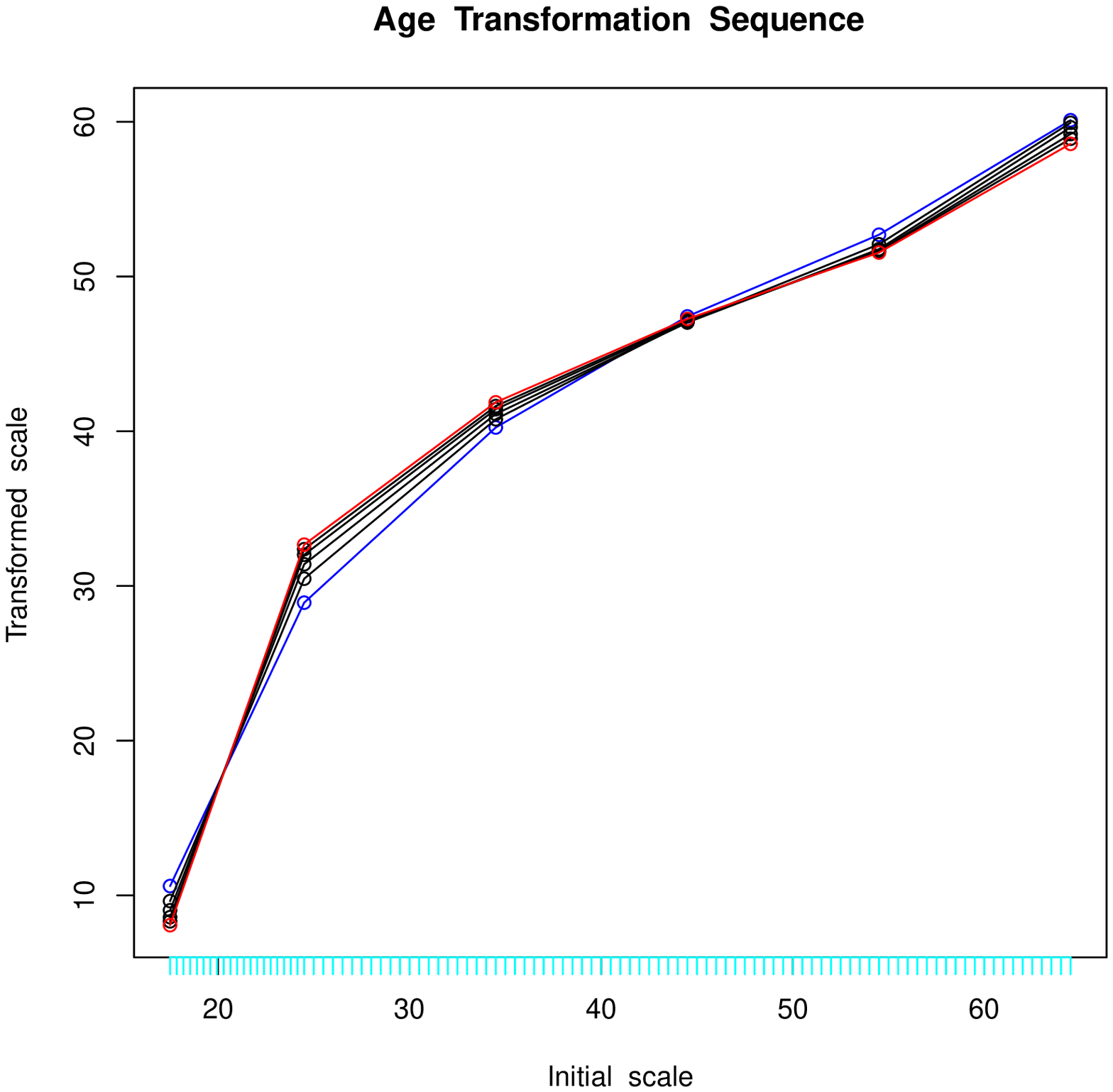}%
\caption{Solutions at successive iterations of the optimal transformation
algorithm on questionnaire data. First is colored blue, two through six black
and seventh red. }%
\label{fig6}%
\end{center}
\end{figure}

Figure \ref{fig6} shows the sequence of transformation solutions for seven
iterations of the optimal transformation procedure described in Section
\ref{opt}. The respective transformations are defined only at the six values
separating the censoring intervals. To aid interpretation lines connecting the
corresponding points representing the same solution are included. Here blue
represents the result of the first iteration, red the last, and black the
intermediate ones. After about four iterations convergence appears to be
achieved. The result at the seventh iteration (red) is chosen as the optimal
transformation estimate $\hat{g}(y)$.%

\begin{figure}
[ptbptb]
\begin{center}
\includegraphics[
height=8.2235in,
width=6.0251in
]%
{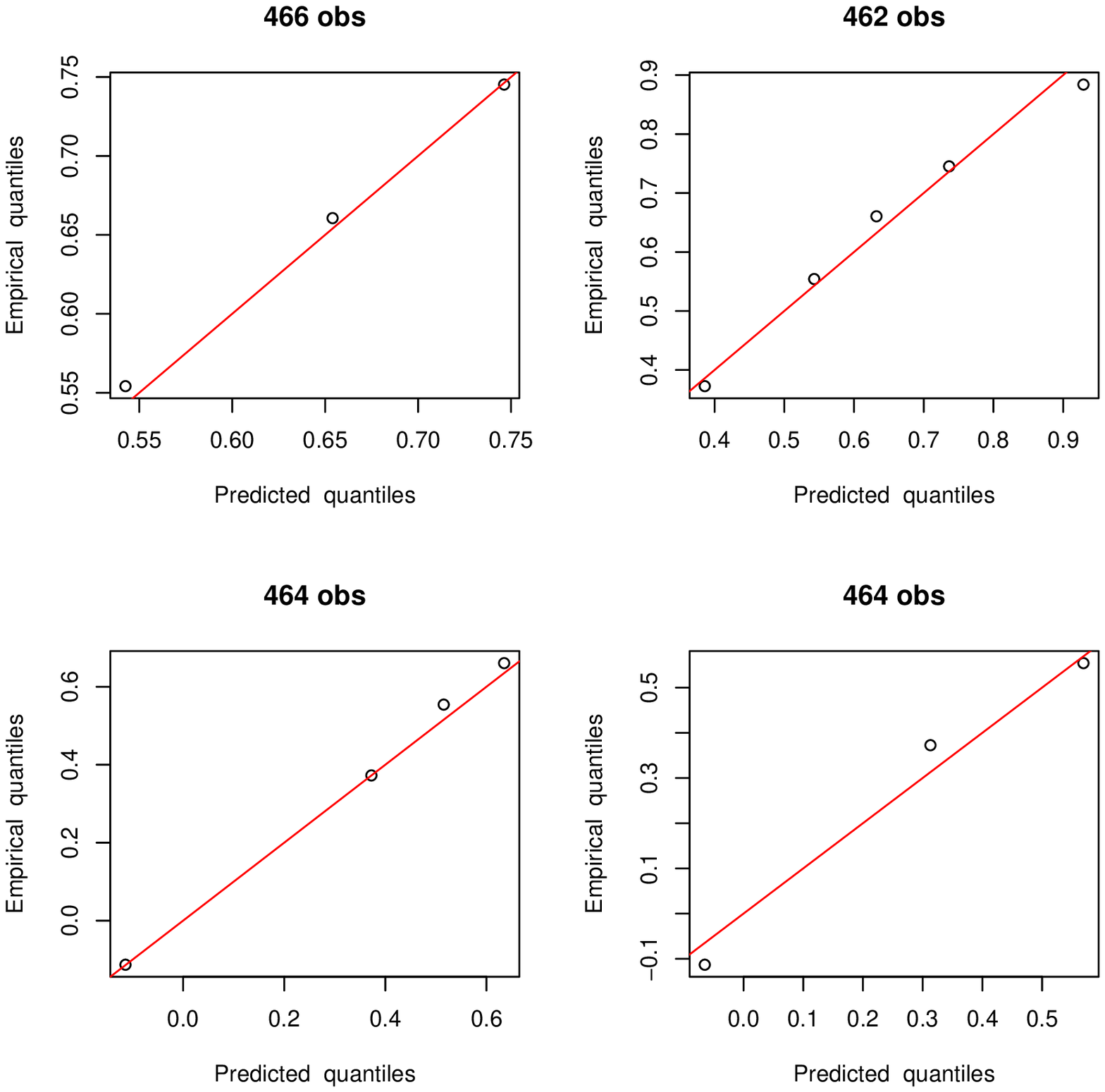}%
\caption{Questionnaire data. Q--Q plots of empirical versus predicted
quantiles for transformed $\hat{g}(y)$ OmniReg solution for four data subsets
obtained by partitioning the corresponding location $\hat{f}(\mathbf{x})$ and
$\hat{s}(\mathbf{x})$ estimates at their respective medians.}%
\label{fig7}%
\end{center}
\end{figure}

Figure \ref{fig7} shows Q--Q plots of the empirical distribution of
\emph{transformed} age $\hat{g}(y)$ versus that predicted by its corresponding
location $\hat{f}(\mathbf{x})$ and scale $\hat{s}(\mathbf{x})$ functions. The
four data subsets are constructed in the same manner as in Fig. \ref{fig5}.
Here in the transformed setting one sees a closer correspondence with model
predictions especially for the larger scale values.%

\begin{figure}
[ptb]
\begin{center}
\includegraphics[
height=7.7435in,
width=6.0563in
]%
{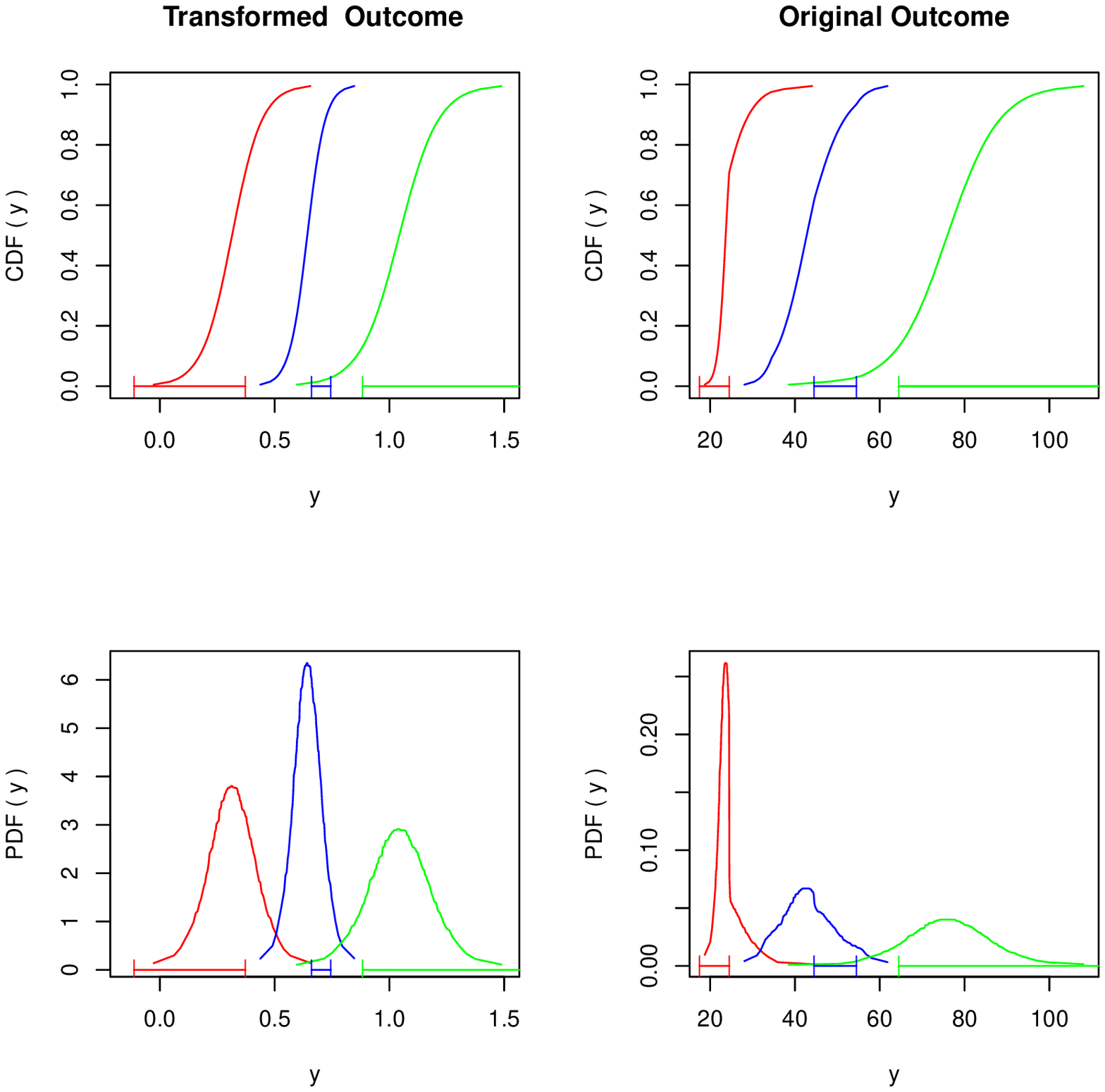}%
\caption{Predicted probability distributions for three questionnaire data test
set observations. Left: transformed setting. Right: original untansformed
(age) setting. Upper: CDF, $\hat{F}(y\,|\,\mathbf{x})$. Lower: PDF, $\hat
{p}(y\,|\,\mathbf{x})$. Bottom intervals represent those specified for the
corresponding three censored observations.}%
\label{fig9}%
\end{center}
\end{figure}
Unlike usual regression procedures that return a single number as a
prediction, OmniReg produces a function representing the predicted
distribution of $y$ given $\mathbf{x}$. Figure \ref{fig9} shows such predicted
functions for $\mathbf{x}$-values of three observations in the validation data
set. The left plots show the distributions in the transformed setting, whereas
the right plots reference the original outcome variable (age). The upper plots
show the cumulative distributions and the lower ones show the corresponding
probability density functions. The intervals indicated at the bottom of each
plot represent the actual censoring intervals for the three chosen
observations (Table 3). Perhaps the most useful representation is the
predicted CDF of the original untransformed $y$ (upper right), since from it
one can directly read probability intervals for $y$ (age) predictions.

\subsection{Online news popularity data\label{mash}}

This data set (Fernandes, Vinagre and P. Cortez 2015) is available for
download from the Irvine Machine Learning Data Repository. It summarizes a
heterogeneous set of features about articles published by Mashable web site
over a period of two years. The goal is to predict the number of shares in
social networks (popularity). There are $N=39797$ observations (articles).
Associated with each are $p=59$ attributes to be used as predictor variables
$\mathbf{x}$. These are described at the download web site
https://archive.ics.uci.edu/ml/datasets/online+news+popularity\#. The outcome
variable for each article is its number of shares. For this analysis these
data are randomly divided into $30000$ observations for model construction,
$5000$ for selecting the number of trees in each model \ (\ref{e5}) (\ref{e6})
, and $4797$ observations for validation.
\begin{figure}
[ptb]
\begin{center}
\includegraphics[
height=3.4809in,
width=5.7536in
]%
{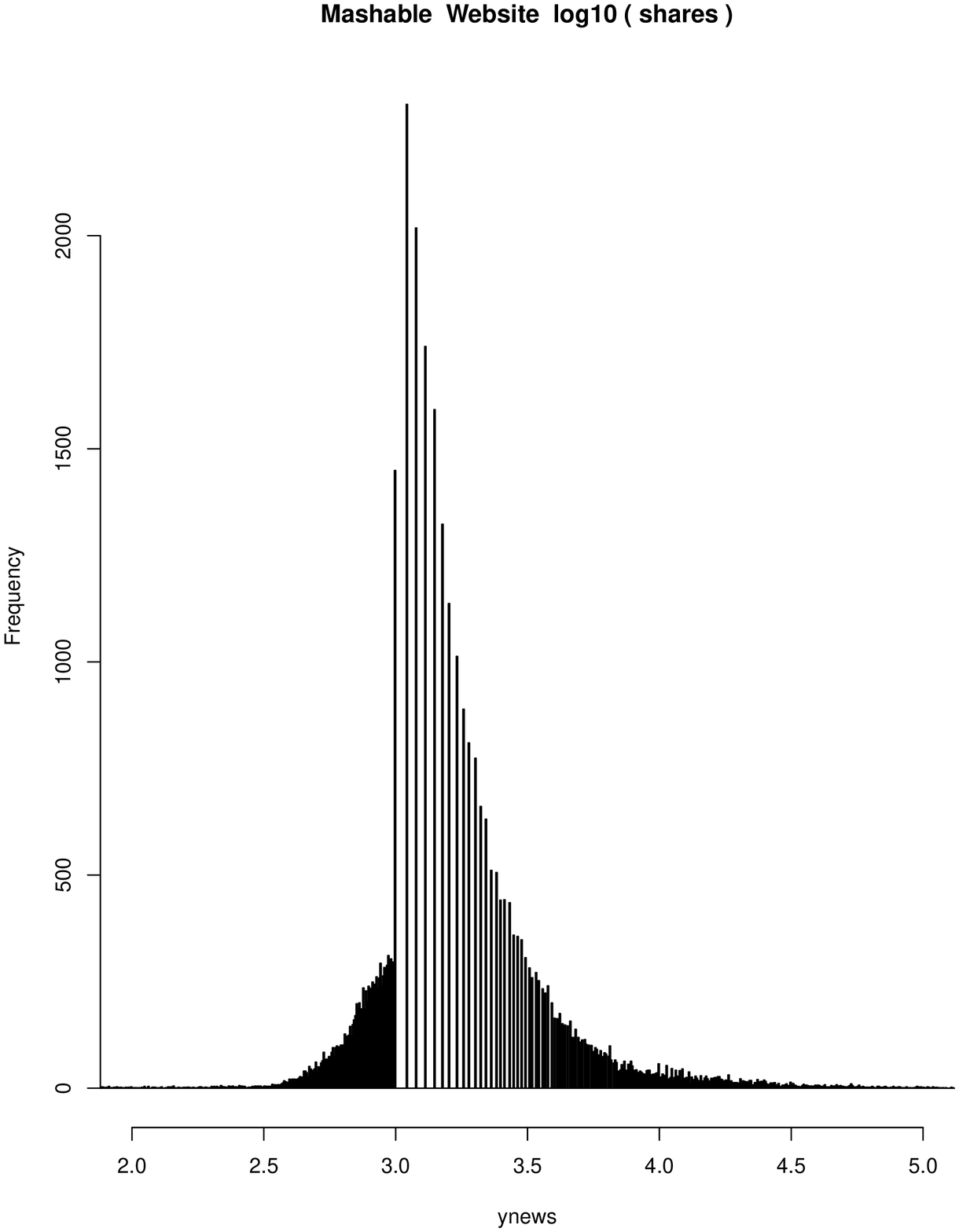}%
\caption{Distribution of $\log_{10}(shares)$ for the online news data.}%
\label{fig10}%
\end{center}
\end{figure}

The number of shares in these data varies from less than 100 to over 100000
and is heavily skewed toward larger values. For data of this type it is common
to apply a log--transform to the outcome variable. Figure \ref{fig10} shows a
histogram of $y=$ $\log_{10}(shares)$ for these data. There seems to be a
sharp change in the nature of this distribution at $1000$ shares. In the first
analysis we directly model $y=$ $\log_{10}(shares)$ without performing any
further transformation (Section \ref{est}). In the second we apply the
technique of Section \ref{opt} to estimate the optimal transformation $\hat
{g}(\log_{10}(shares))$.

\begin{figure}
[ptb]
\begin{center}
\includegraphics[
height=7.7531in,
width=6.0347in
]%
{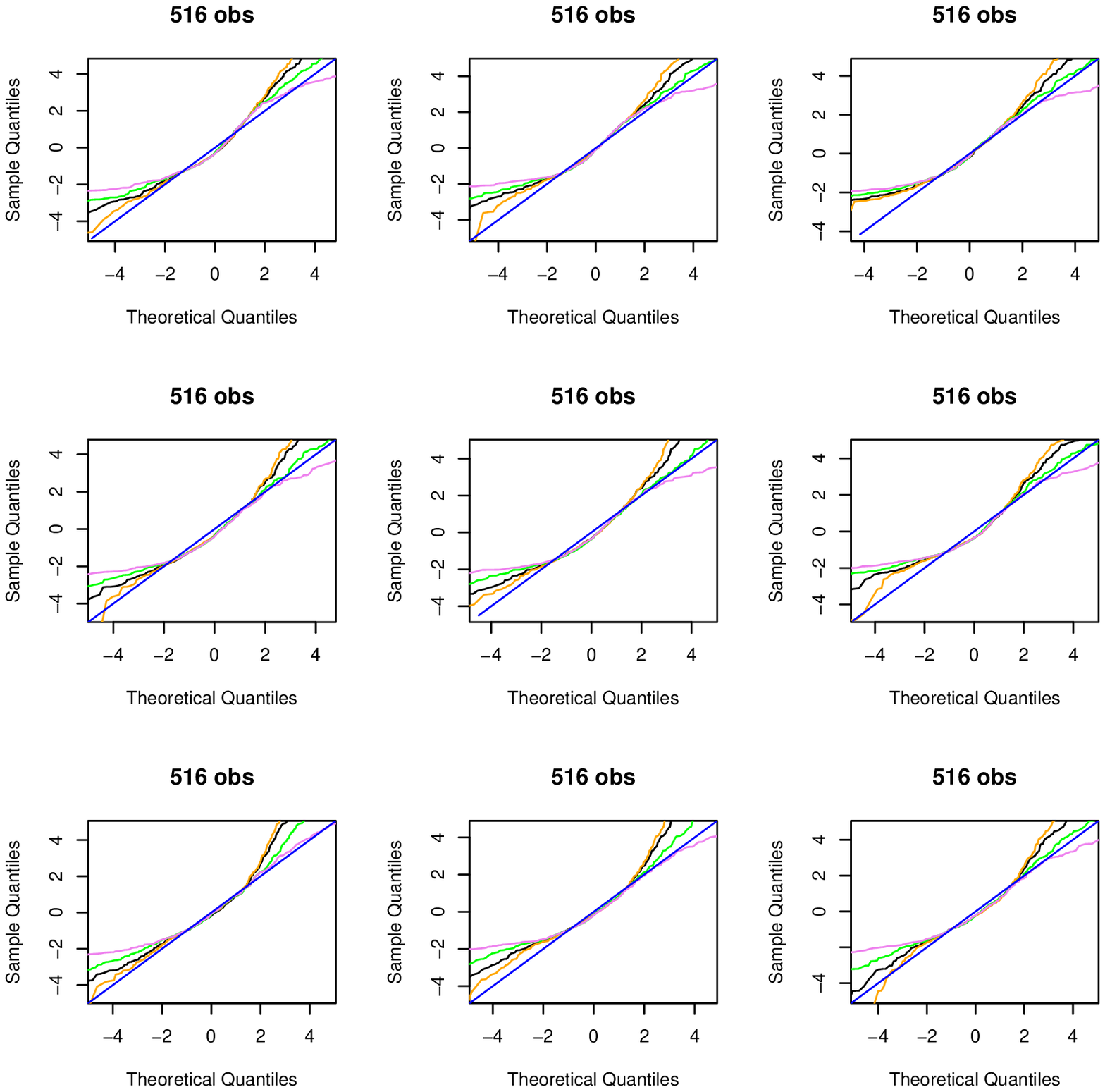}%
\caption{Online news popularity data. Diagnostic Q-Q plots of actual vs.
predicted distribution of untransformed standardized residuals $(y\,\,-$
$\hat{f}(\mathbf{x}))/\hat{s}(\mathbf{x})$ for nine data subsets delineated by
joint intervals of the $1/3$ quantiles of estimated location $\hat
{f}(\mathbf{x})$ and scale $\hat{s}(\mathbf{x})$. The black, orange, green and
violet lines respectively represent comparisons to logistic, normal, Laplace
and slash distributions.}%
\label{fig11}%
\end{center}
\end{figure}

Since these data are uncensored one can directly analyze the predicted
residuals as described in Section \ref{stdres}. Fig. \ref{fig11} shows nine
standardized residual Q--Q plots based on the untransformed model using nine
different subsets of the validation data set. These subsets were constructed
as described in Section \ref{quest} (Fig. \ref{fig5}), but with partitions at
the corresponding 33\% quantiles of $\hat{f}(\mathbf{x}_{i})$ and $\hat
{s}(\mathbf{x}_{i})$ to create nine regions.

Each frame in Fig. \ref{fig11} \ displays four Q-Q plots. The black lines
represent a comparison of the estimated standardized residuals to a logistic
distribution. For reference, the orange, green and violet lines show
comparisons to the normal, Laplace and slash distributions respectively. The
slash (Rogers and Tukey 1972) is the distribution of the ratio of normal and
uniform random variables and has very heavy tails. The blue line in each plot
is the 45--degree diagonal. Here one sees that the standardized residuals
predicted by the model do not closely follow a logistic distribution (black
line) for any of the location--scale defined data subsets. This suggests that
inference based on this model using the log--transformation is highly suspect.%

\begin{figure}
[ptb]
\begin{center}
\includegraphics[
height=3.6331in,
width=5.0246in
]%
{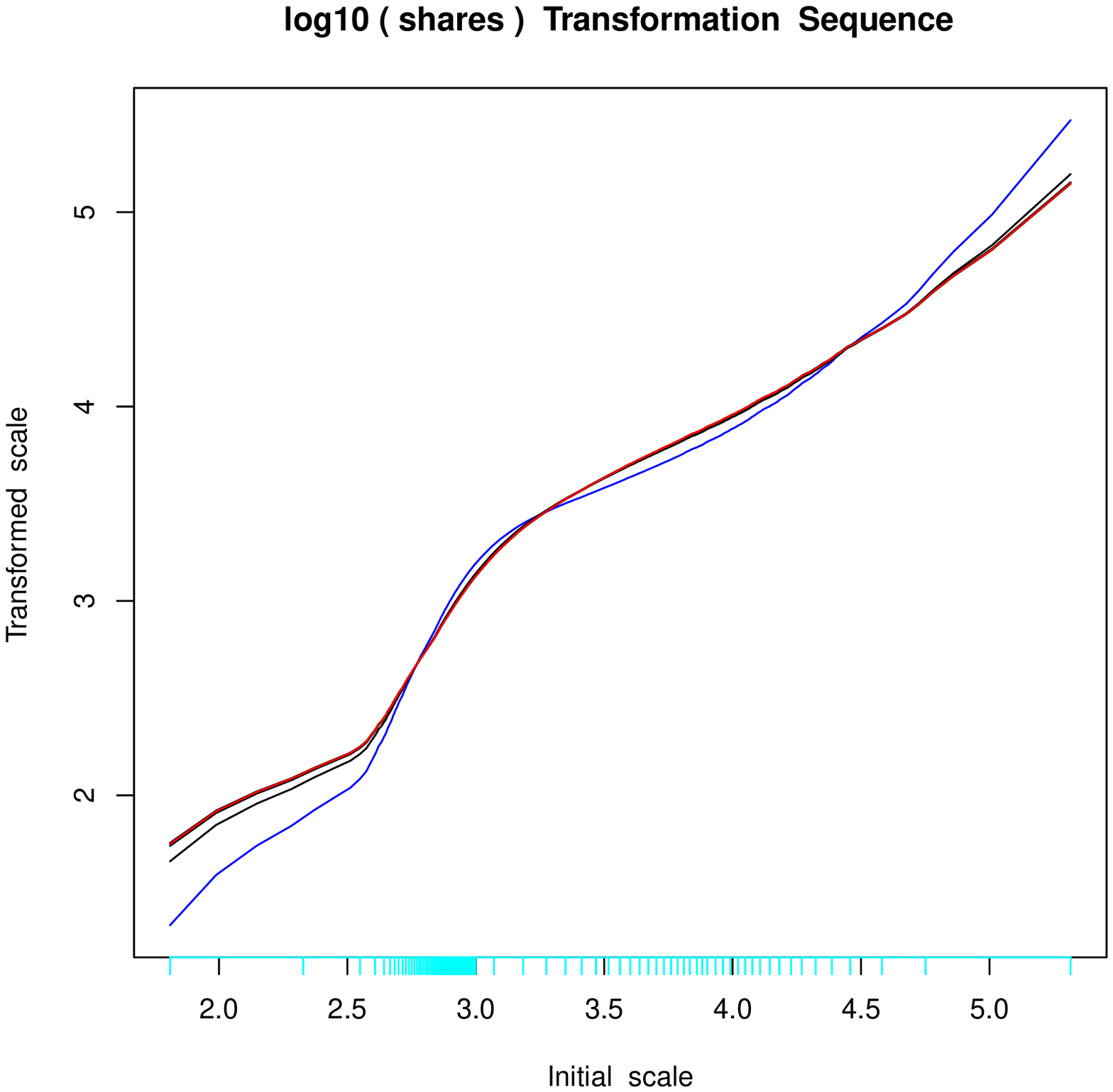}%
\caption{Solutions at successive iterations of the optimal transformation
algorithm on online news popularity data. First is colored blue, two through
six black and seventh red. The blue hash marks below the abscissa delineate
1\% intervals \ of the $y$-distribution.}%
\label{fig12}%
\end{center}
\end{figure}

Applying the optimal transformation procedure of Section \ref{opt} to the
log-transformed data one obtains the sequence of transformation estimates for
each iteration shown in Fig. \ref{fig12}. The first estimate is colored blue,
the seventh red, and the intermediates black. The blue hash marks below the
abscissa delineate 1\% intervals \ of the $y$-distribution. Convergence
appears to occur after three iterations.%

\begin{figure}
[ptb]
\begin{center}
\includegraphics[
height=8.1699in,
width=6.0658in
]%
{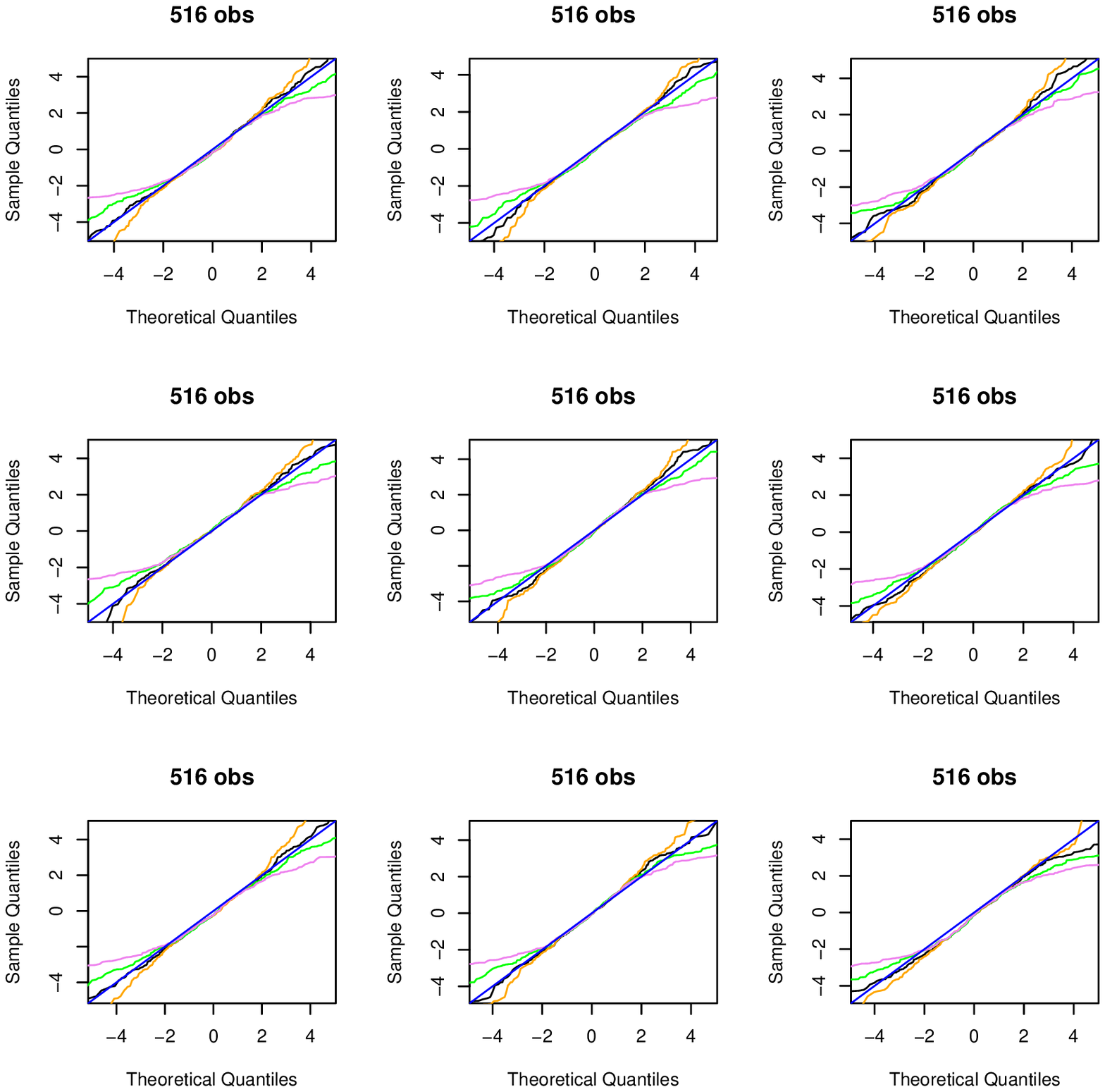}%
\caption{Online news popularity data. Diagnostic Q-Q plots of actual vs.
predicted distribution of optimal transformed standardized residuals $(\hat
{g}(y)\,-$ $\hat{f}(\mathbf{x}))/\hat{s}(\mathbf{x})$ for nine data subsets
delineated by joint intervals of the $1/3$ quantiles of estimated location
$\hat{f}(\mathbf{x})$ and scale $\hat{s}(\mathbf{x})$. The black, orange,
green and violet lines respectively represent comparisons to logistic, normal,
Laplace and slash distributions.}%
\label{fig13}%
\end{center}
\end{figure}

Using the model $\hat{g}(\log_{10}(shares))$ based on the seventh (red)
transformation produces the corresponding standardized residual plots shown in
\ Fig. \ref{fig13}. Here the predicted residuals very closely follow a
standard logistic distribution (black line) for all data subsets. A slight
exception might be the lower right plot (small location, large scale) where
the extreme positive residuals appear somewhat too large. This diagnostic
provides little evidence of overall lack of fit of the transformed model or
violation of its\ assumptions. Of course, one cannot absolutely exclude the
possibility that there are other diagnostics that might demonstrate such evidence.%

\begin{figure}
[ptb]
\begin{center}
\includegraphics[
height=7.9943in,
width=6.0044in
]%
{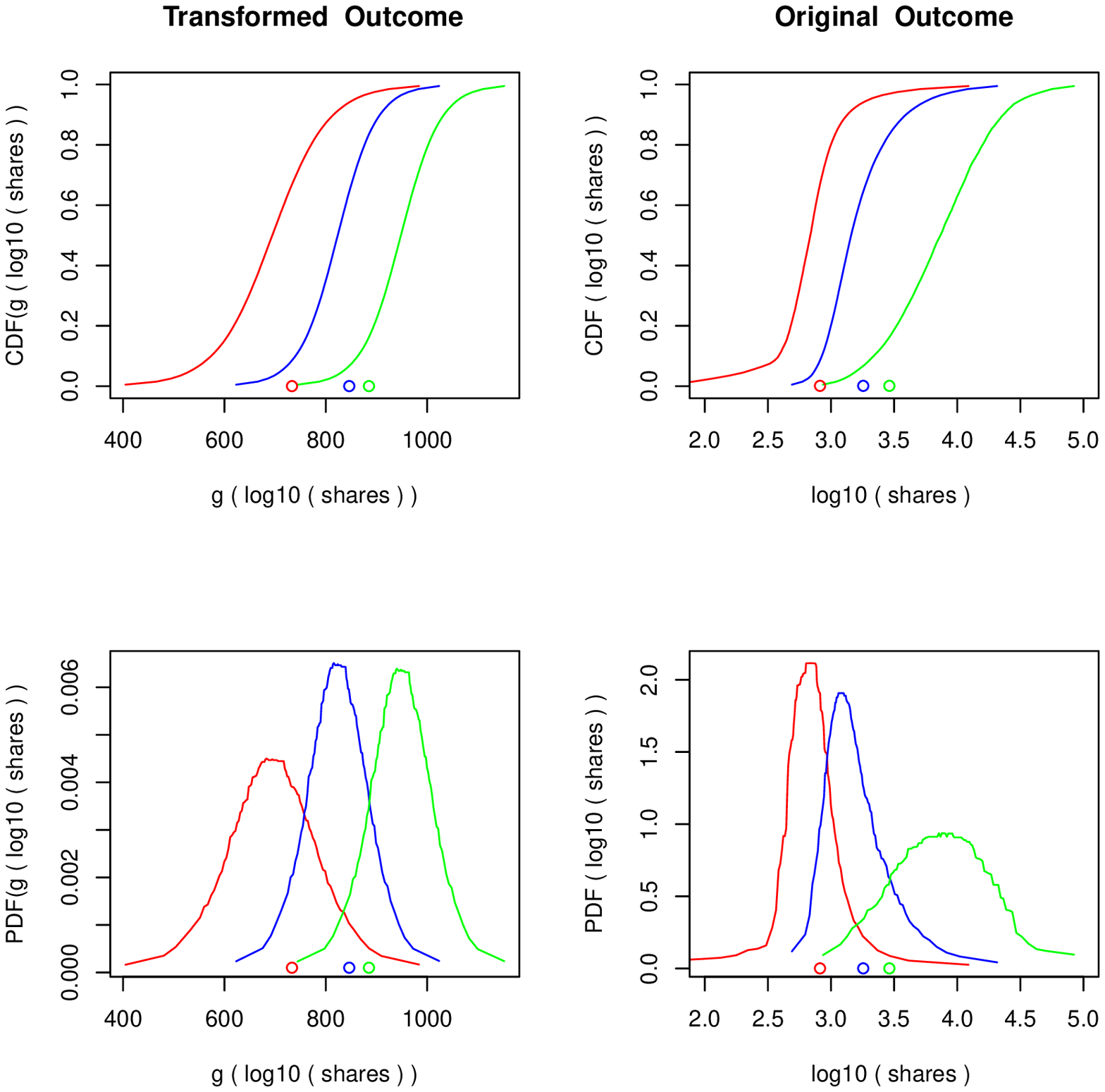}%
\caption{Predicted probability distributions for three test set observations
from the online news popularity data. Left: transformed setting. Right:
original untansformed (log--shares) setting. Upper: CDF, $\hat{F}%
(y\,|\,\mathbf{x})$. Lower: PDF, $\hat{p}(y\,|\,\mathbf{x})$. Bottom points
represent the corresponding recorded $\log_{10}(shares)$ for the three
selected observations.}%
\label{fig14}%
\end{center}
\end{figure}

Figure \ref{fig14} shows curves representing the model predicted probability
distributions of $\log_{10}(shares)$ for $\mathbf{x}$-values of three
observations taken from the validation data set. The format is the same as in
Fig. \ref{fig9}. The circles at the bottom of each plot represent the actual
realized $\log_{10}(shares)$ for these three chosen observations. Distribution
asymmetry and heteroscedasticity are evident in the untransformed setting
(right frames). Prediction probability intervals can be directly determined
from the upper right plot.%

\subsection{Million Song Data Set\label{mds}}

These data are a subset of the Million Song Data Set and are also available
for download from the Irvine Machine Learning Data Repository. It consists of
a training data set of 463715 and a test set of 51630 song recordings. We
divide a randomly selected subsample of the training data into a learning data
subset of 50000 observations and one for model selection consisting of \ 20000
observations. All diagnostics are performed on the 51630 song test data set.
There are 89 predictor variables measuring various acoustic properties of the
recordings. The outcome variable is the year the recording was made\ ranging
from 1922 to 2011.%

\begin{figure}
[ptb]
\begin{center}
\includegraphics[
height=3.0874in,
width=4.9208in
]%
{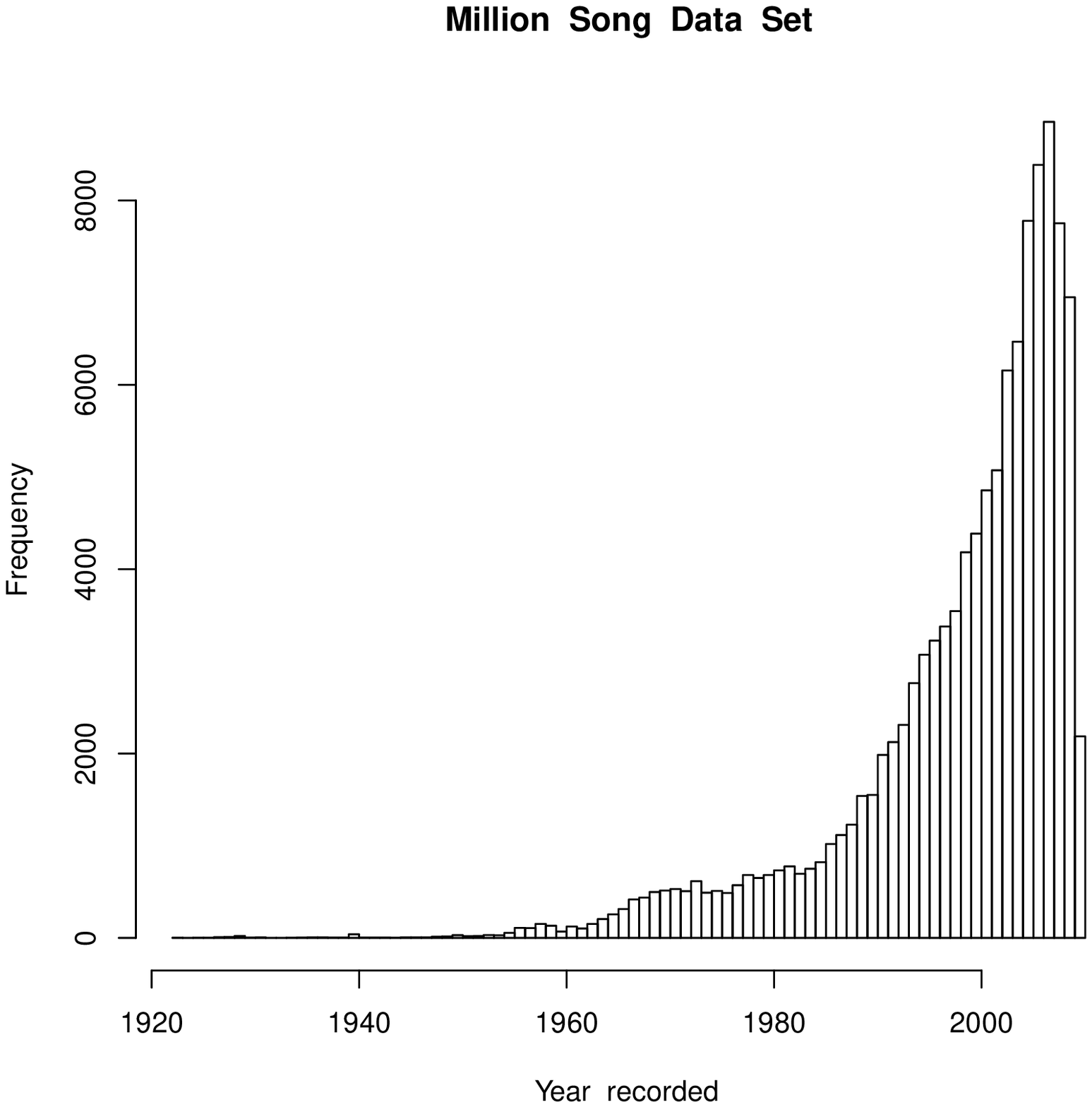}%
\caption{Distribution of recording year for the million song data set.}%
\label{fig15}%
\end{center}
\end{figure}

\begin{figure}
[ptb]
\begin{center}
\includegraphics[
height=7.5559in,
width=6.0243in
]%
{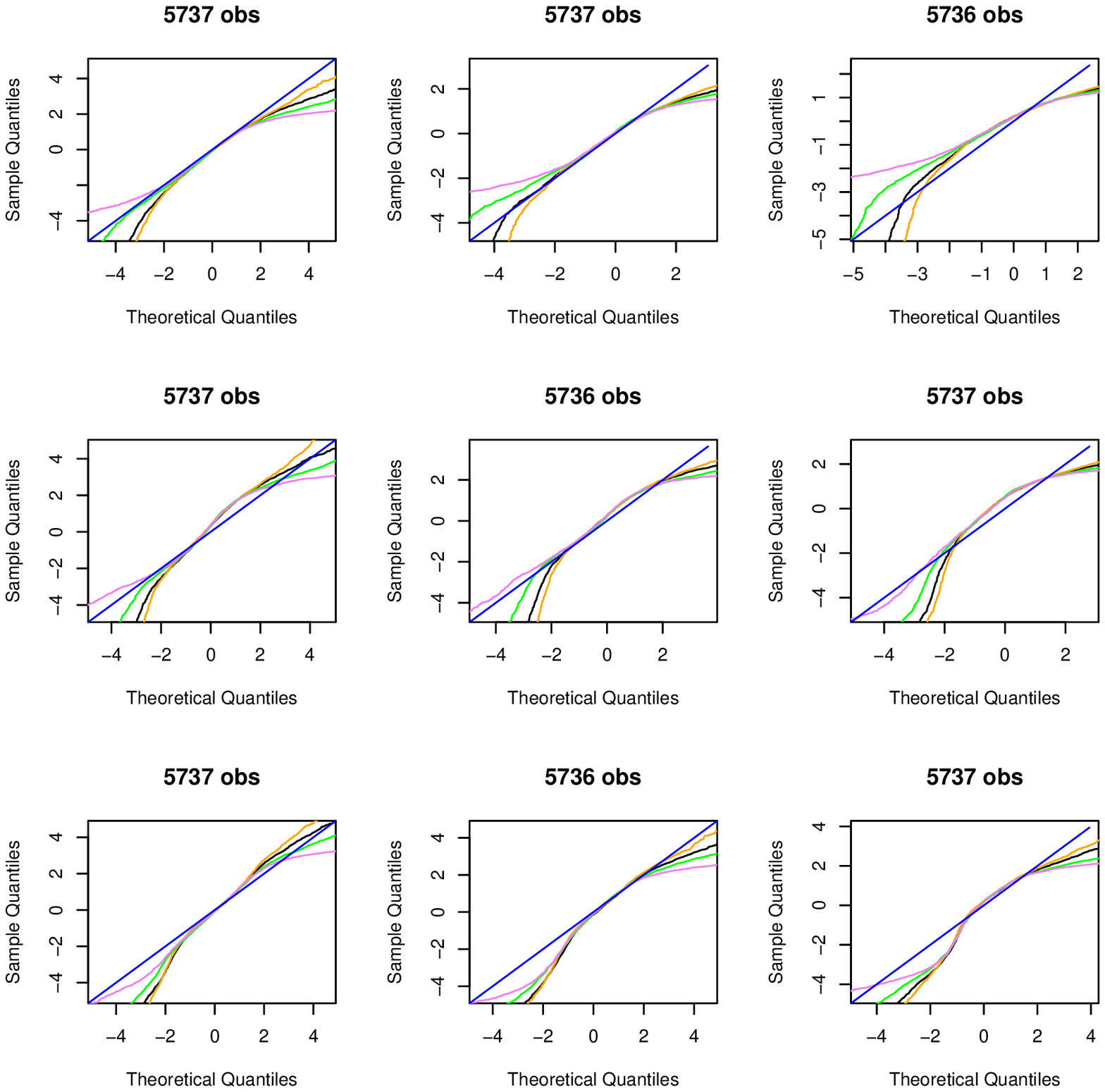}%
\caption{Million song data set. Diagnostic Q-Q plots of actual vs. predicted
distribution of untransformed standardized residuals $(y\,-$ $\hat
{f}(\mathbf{x}))/\hat{s}(\mathbf{x})$ for nine data subsets delineated by
joint intervals of the $1/3$ quantiles of estimated location $\hat
{f}(\mathbf{x})$ and scale $\hat{s}(\mathbf{x})$. The black, orange, green and
violet lines respectively represent comparisons to logistic, normal, Laplace
and slash distributions.}%
\label{fig16}%
\end{center}
\end{figure}

Figure \ref{fig15} shows a histogram of the outcome variable $y$. There are
relatively few songs in the data set recorded before 1955. Modeling the data
without transformations (Section \ref{est}) produces the standardized residual
diagnostic plots on the test data set shown in Fig. \ref{fig16}. The data
subsets are constructed in the same manner as for Fig. \ref{fig11}. Again,
none of the predicted standardized residual distributions in any of the data
subsets are remotely close to being standard logistic (black line).%

%

\begin{figure}
[ptb]
\begin{center}
\includegraphics[
height=3.3892in,
width=4.4417in
]%
{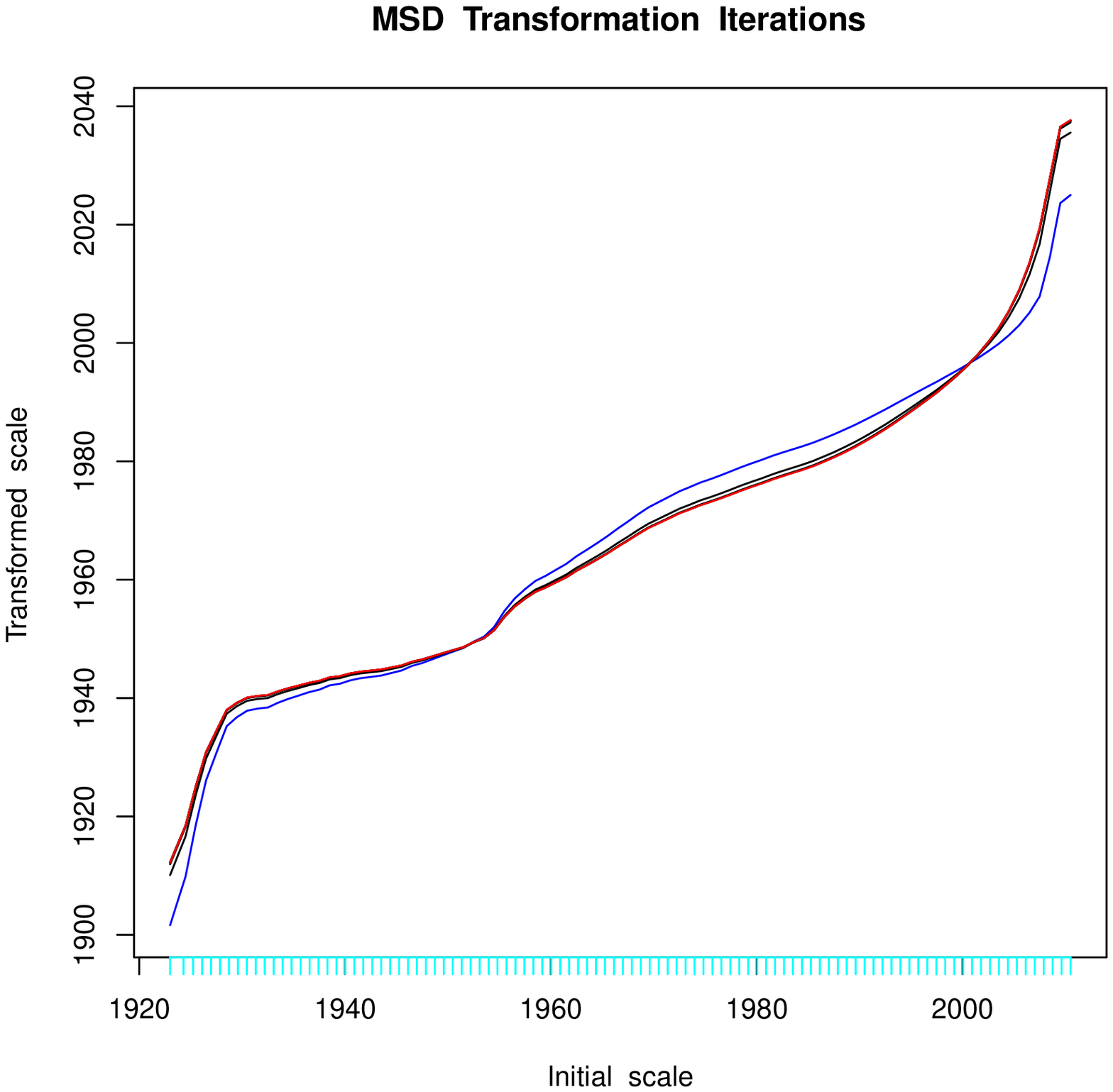}%
\caption{Solutions at successive iterations of the optimal transformation
algorithm on million song data set. First is colored blue, two through six
black and seventh red. }%
\label{fig17}%
\end{center}
\end{figure}
%

\begin{figure}
[ptb]
\begin{center}
\includegraphics[
height=7.8793in,
width=6.0139in
]%
{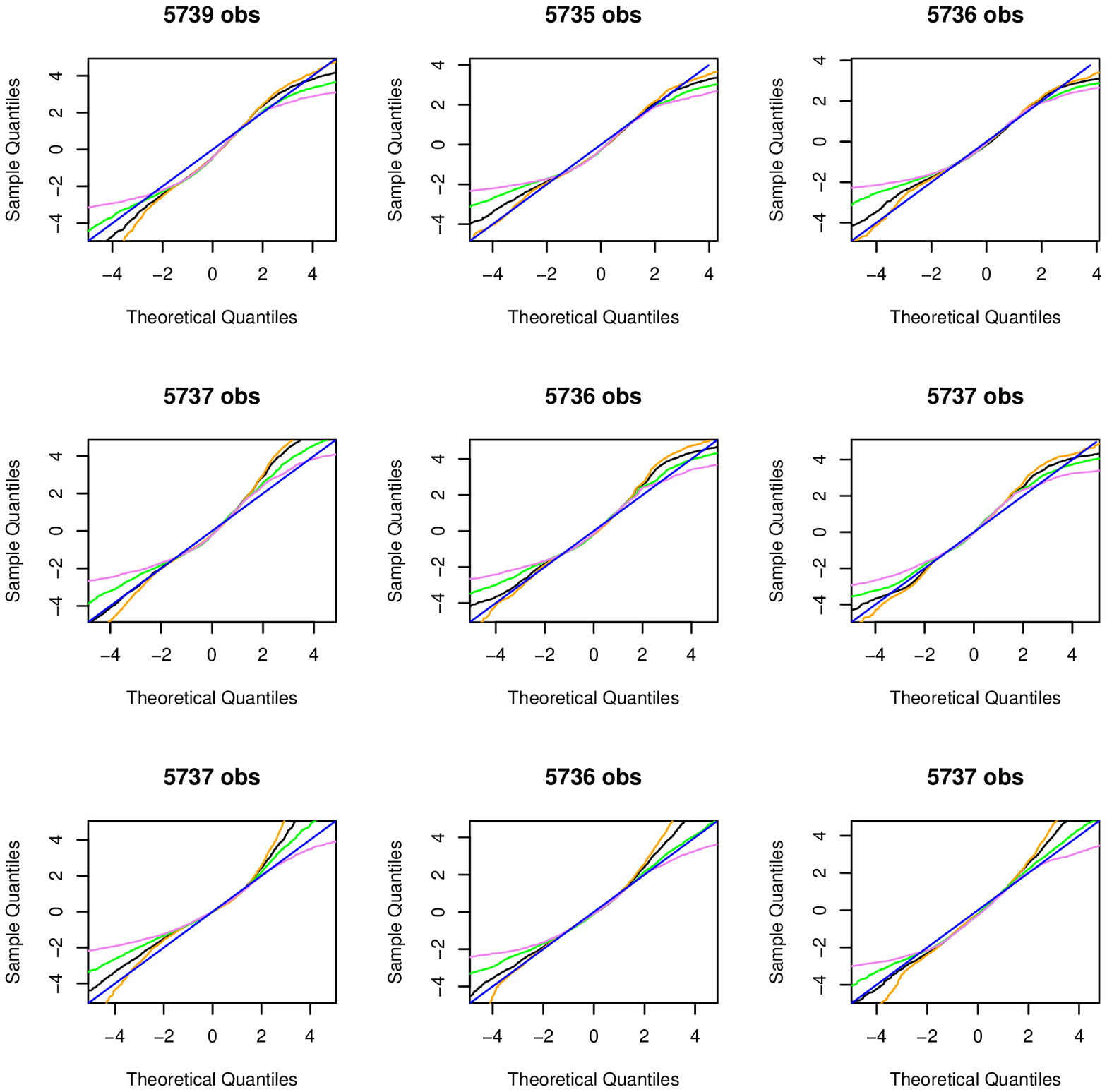}%
\caption{Million song data set. Diagnostic Q-Q plots of actual vs. predicted
distribution of transformed standardized residuals $(\hat{g}(y)\,\,-$ $\hat
{f}(\mathbf{x}))/\hat{s}(\mathbf{x})$ for nine data subsets delineated by
joint intervals of the $1/3$ quantiles of estimated location $\hat
{f}(\mathbf{x})$ and scale $\hat{s}(\mathbf{x})$. The black, orange, green and
violet lines respectively represent comparisons to logistic, normal, Laplace
and slash distributions.}%
\label{fig18}%
\end{center}
\end{figure}

Applying the optimal transformation strategy of Section \ref{opt} produces the
sequence of transformation estimates for\ seven iterations shown in Fig.
\ref{fig17}. There is little change after two iterations. Figure \ref{fig18}
shows the corresponding standardized residual Q--Q plots for the seventh
transformed solution. Although not perfect, these residuals much more closely
follow a logistic distribution (black line) than for the untransformed
solution (Fig. \ref{fig16}). The residual distributions for some of the
samples are seen to differ mainly for extreme positive values where the data
have somewhat narrower tails.

\begin{figure}
[ptb]
\begin{center}
\includegraphics[
height=7.9511in,
width=6.0243in
]%
{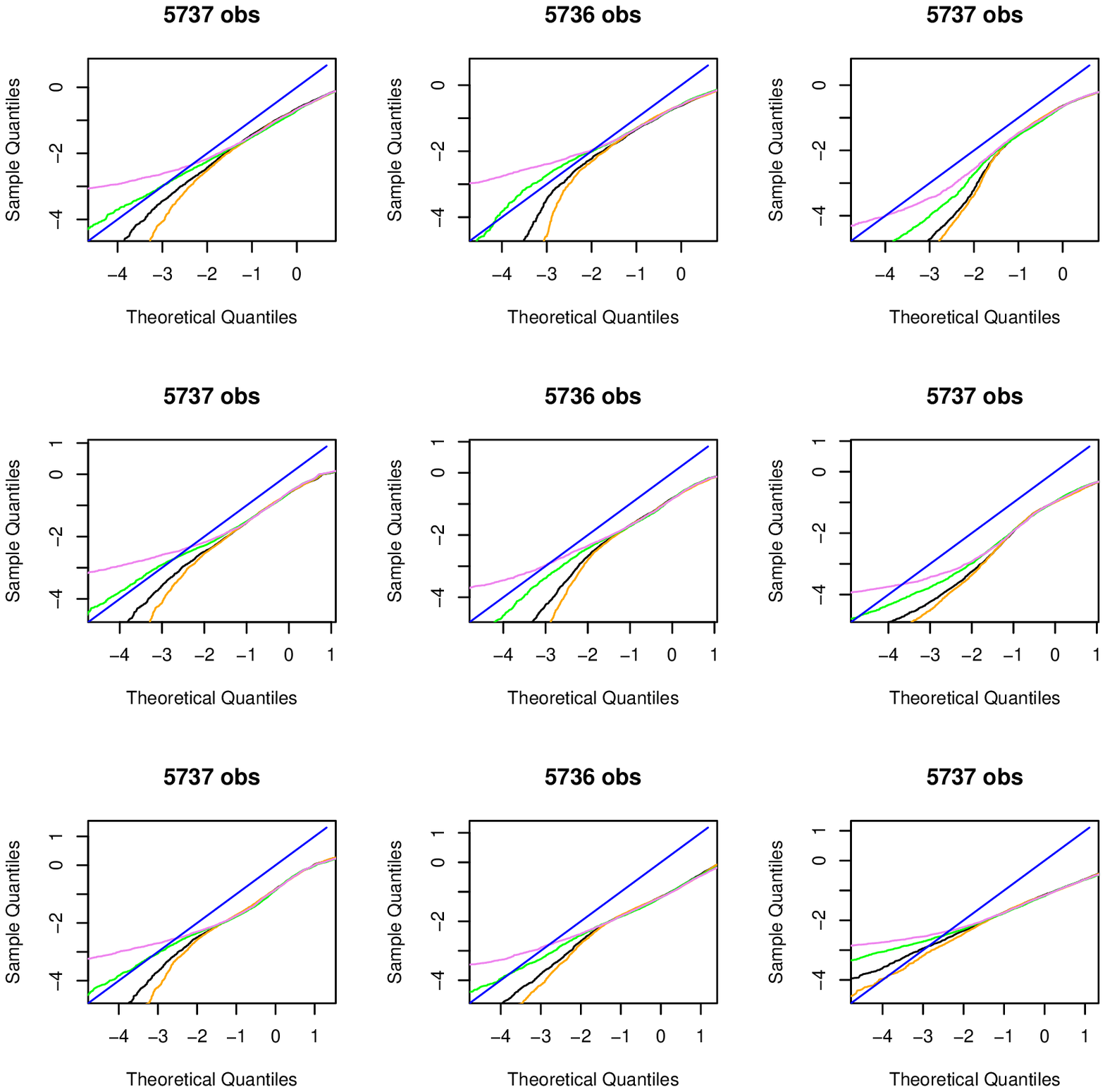}%
\caption{Million song data set. Diagnostic Q-Q plots of actual vs. predicted
distribution of untransformed asymmetric standardized residuals (22) for nine
data subsets delineated by joint intervals of the $1/3$ quantiles of estimated
location $\hat{f}(\mathbf{x})$ and those of the geometric mean $\sqrt{\hat
{s}_{l}(\mathbf{x})\hat{s}_{u}(\mathbf{x})}$ of the scales. The black, orange,
green and violet lines respectively represent comparisons to logistic, normal,
Laplace and slash distributions.}%
\label{fig19}%
\end{center}
\end{figure}

Figure \ref{fig18} suggests that the optimal transformation strategy of
Section \ref{opt} based on the additive symmetric error model (\ref{e10.5})
has not produced totally accurate probability estimates. This in turn suggests
trying the asymmetric procedure of Section \ref{asym}. Figure \ref{fig19}
shows the resulting diagnostic Q--Q plots when the second asymmetric algorithm
of Section \ref{asym} is applied to the original \emph{untransformed} outcome
data. \ The result is seen to be (at best) no better than the corresponding
symmetric error results (Fig. \ref{fig16}).%

\begin{figure}
[ptb]
\begin{center}
\includegraphics[
height=3.467in,
width=4.6855in
]%
{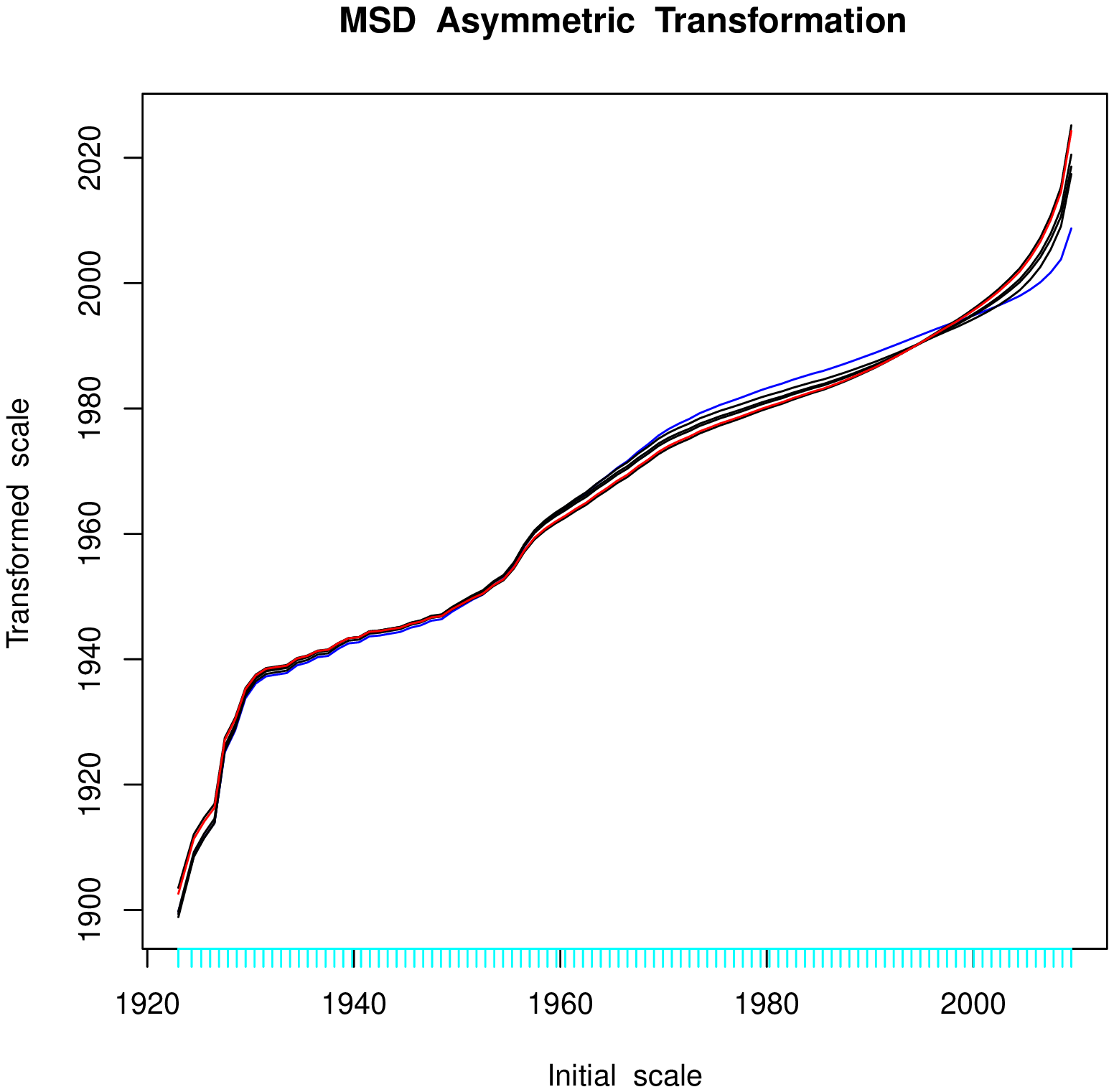}%
\caption{Solutions at successive iterations of the asymmetric optimal
transformation algorithm on million song data set. First is colored blue, two
through six black and seventh red. }%
\label{fig30}%
\end{center}
\end{figure}

Figure \ref{fig30} shows the sequence of transformations produced by the
optimal transformation strategy of Section \ref{opt} used in conjunction with
the asymmetric estimation procedure of Section \ref{asym}. The estimated
optimal transformation (red) has the same general shape as the one produced by
the symmetric procedure (Fig. \ref{fig17}) with small to moderate differences
mainly at the lower extreme.%

\begin{figure}
[ptb]
\begin{center}
\includegraphics[
height=7.8421in,
width=6.0399in
]%
{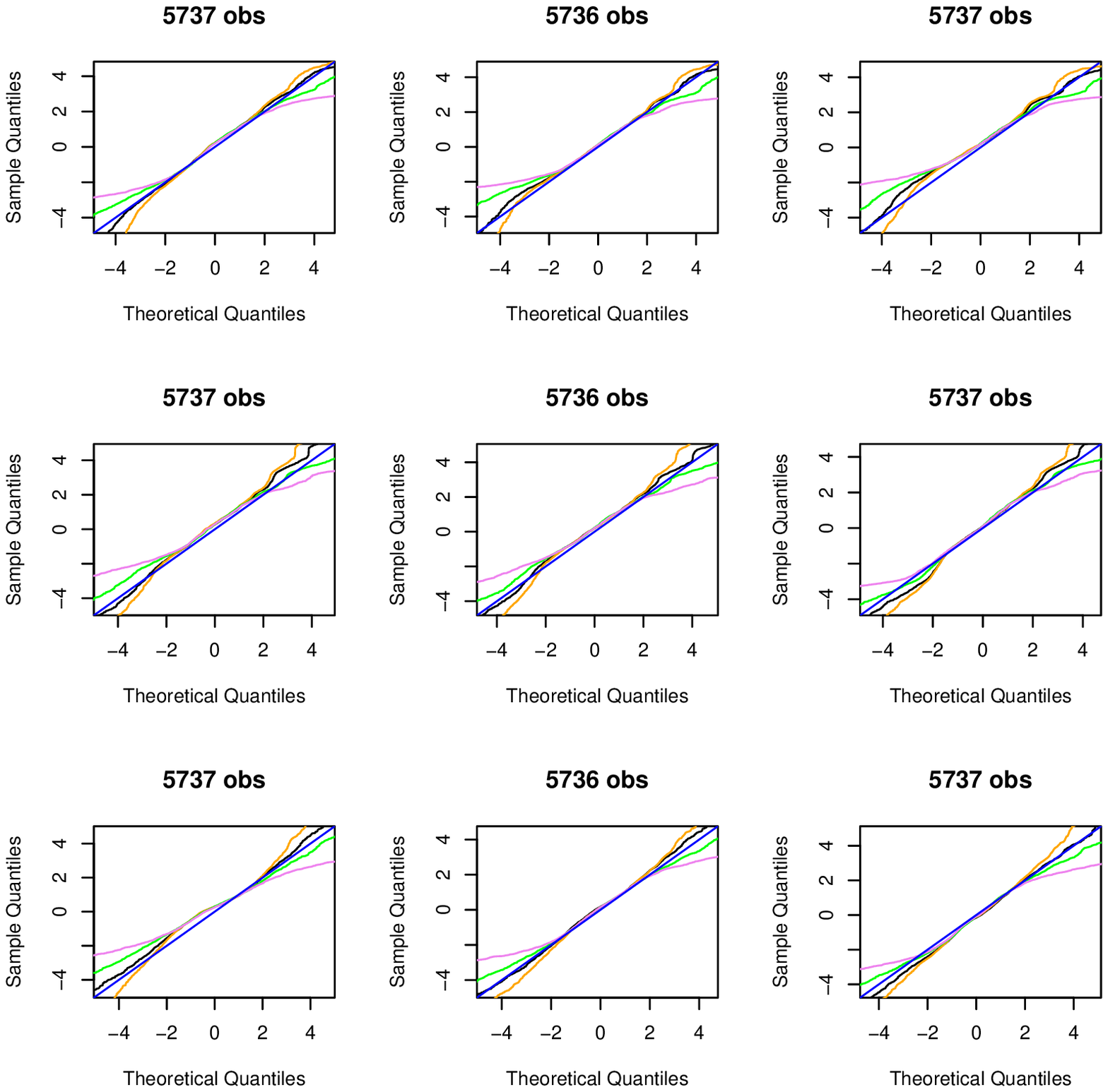}%
\caption{Million song data set. Diagnostic Q-Q plots of actual vs. predicted
distribution of transformed asymmetric standardized residuals (22) for nine
data subsets delineated by joint intervals of the $1/3$ quantiles of estimated
location $\hat{f}(\mathbf{x})$ and geometric mean $\sqrt{\hat{s}%
_{l}(\mathbf{x})\hat{s}_{u}(\mathbf{x})}$ of the scales estimates. The black,
orange, green and violet lines respectively represent comparisons to logistic,
normal, Laplace and slash distributions.}%
\label{fig31}%
\end{center}
\end{figure}

Figure \ref{fig31} shows the resulting diagnostic plots produced by the
asymmetric error model in its optimally transformed setting. For this
diagnostic the respective data subsets represent the validation data
partitioned at the 33\% quantiles of their location (mode) estimates $\hat
{f}(\mathbf{x})$ (bottom to top) and the 33\% quantiles of the geometric mean
of their two scale estimates $\sqrt{\hat{s}_{l}(\mathbf{x})\,\hat{s}%
_{u}(\mathbf{x})}$ (left to right). The (asymmetric) standardized residuals
(\ref{e35}) are seen to very closely follow a standard logistic distribution
(\ref{e2}) for all data subsets.%

\begin{figure}
[ptb]
\begin{center}
\includegraphics[
height=3.6391in,
width=5.0566in
]%
{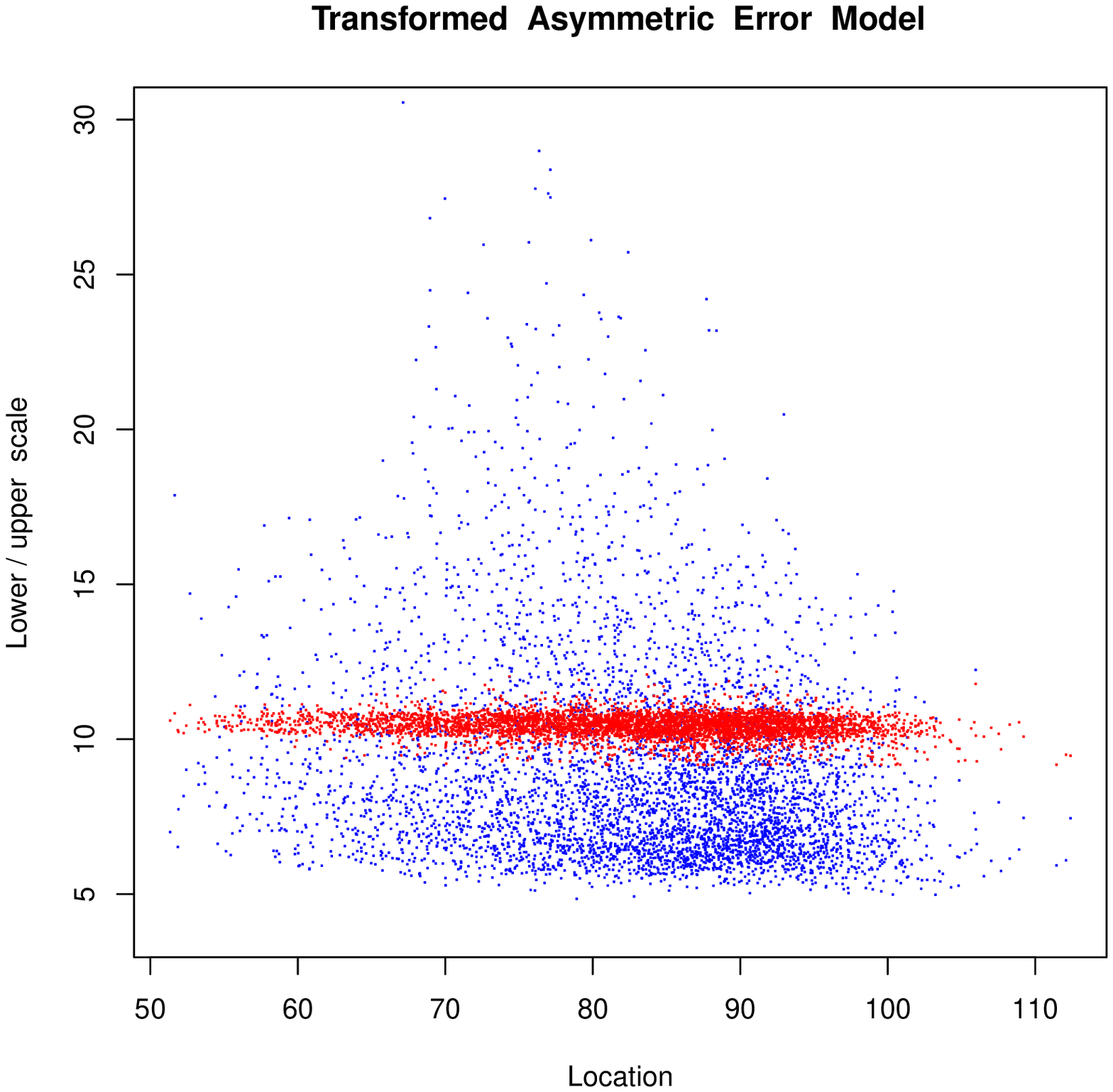}%
\caption{Million song data. Upper scale estimates $\hat{s}_{u}(\mathbf{x})$
(red) and lower scale estimates $\hat{s}_{l}(\mathbf{x})$ (blue) vs. mode
$\hat{f}(\mathbf{x})$ for asymmetric optimal transformed setting.}%
\label{fig32}%
\end{center}
\end{figure}

Figure \ref{fig32} shows plots of the lower $\hat{s}_{l}(\mathbf{x})$ (blue)
and upper $\hat{s}_{u}(\mathbf{x})$ (red) scale estimates against
corresponding location estimates $\hat{f}(\mathbf{x})$ in the optimal
transformed setting. Here the upper scales are seen to be almost constant
(homoscedastic) whereas the lower scales vary by roughly a factor of four.
Neither seem to have any association with the location estimates.
%

\begin{figure}
[ptb]
\begin{center}
\includegraphics[
height=7.9891in,
width=6.0399in
]%
{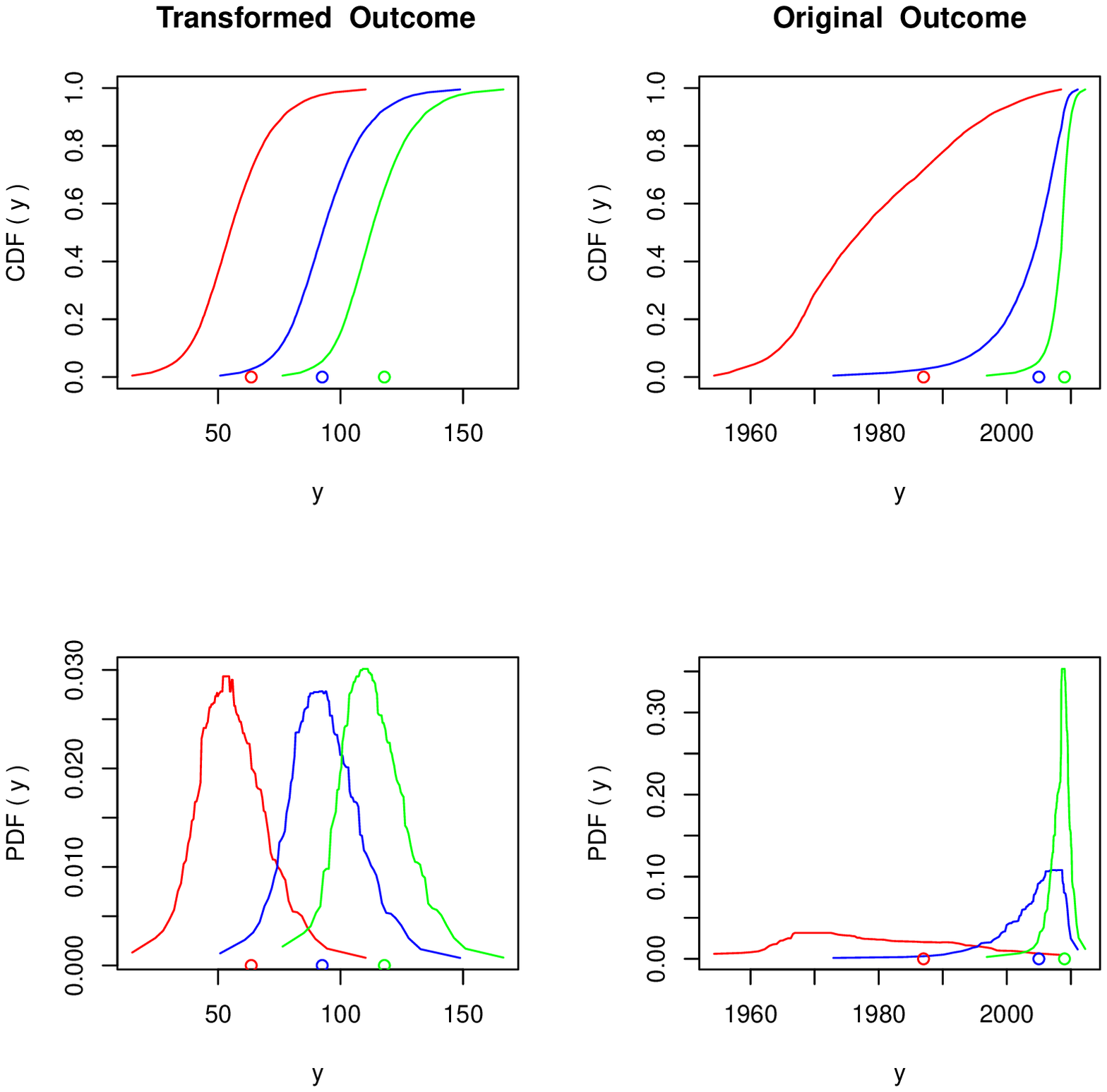}%
\caption{Predicted probability distributions based on the optimally
transformed asymmetric model for three test set observations from the million
song data. Left: transformed setting. Right: original untansformed (year)
setting. Upper: CDF, $\hat{F}(y\,|\,\mathbf{x})$. Lower: PDF, $\hat
{p}(y\,|\,\mathbf{x})$. Bottom points represent the corresponding dates for
the three selected recordings.}%
\label{fig33}%
\end{center}
\end{figure}

Figure \ref{fig33} shows the predicted recording year probability
distributions for three recordings in the test data set based on the
transformed asymmetric model. The format is the same as that in Figs.
\ref{fig9} and \ref{fig14}. Unlike the distributions in those figures that are
based on symmetric transformed models, here one sees some asymmetry of the
distributions in the transformed setting. In the \emph{original} outcome
setting (recording year) there is seen to be massive skewness and
heteroscedasticity. The predictive model is pretty certain that the recording
indicated by green was made after 2005. For the blue recording it predicts
somewhere between 1990 and 2010. It reports having little indication as to
when the recording in red was made except that it is likely, but not
certainly, before the other two. The circles at the bottom of each plot
represent the actual dates for these three recordings.

\section{Ordinal regression\label{ord}}

In the examples of Section \ref{examp} there is a known measurement scale for
the values of the outcome variable $y$. Values of $y$ as well as the interval
bounds $\{a_{i},b_{i}\}_{1}^{N}$ in (\ref{e3}) (\ref{e4}) reference this
scale. In ordinal regression no such measurement scale exists. Only a relative
order relation among the $y$-values is specified. Usually ordinal regression
is applied in the context of many ties so that there are only a few $K$ unique
grouped values such as $y\in\{$\emph{small, medium, large, extra large}$\}$.
For any set of joint predictor variable values $\mathbf{x}$ the main goal is
to predict $p(y\in k\,|\,\mathbf{x})$ where $k=1\cdot\cdot\cdot K$ labels the
groups. In this sense ordinal regression can be viewed as classification where
there is an order relation among the class labels.

OmniReg with optimal transformations (Sections \ref{est} , \ref{opt} and
\ref{asym}) is fundamentally a (generalized) ordinal regression procedure. Its
solutions depend only on the values of the cumulative distribution (ranks) of
$y$, $\hat{F}(y)$ (\ref{e10.9}). As noted in Section \ref{data} tied
$y$-values can be considered as interval censored with lower bound at the
midpoint of its value and the next lower value, and upper bound the midpoint
of its value and the next higher value. The lower bound of the first group can
be taken to be $-\infty$ and the upper bound of the last $\infty$. This was
the strategy used in Section \ref{quest} (Table 3).

Note that for the problem analyzed in Section \ref{quest} there was no
fundamental requirement that the censoring intervals be on the age measurement
scale. Any scale that produces the same order (grouping) could have been used.
Changing the measurement scale is equivalent to changing the starting
transformation for the iterative procedure of Section \ref{opt}. Except for
possible numerical effects the resulting solutions should be essentially equivalent.

Censoring in ordinal regression is easily incorporated with this approach. If
it is unknown which of several adjacent groups contains training observation
$y_{i}$, its lower bound $a_{i}$ is set to the minimum of the lower bounds
over those groups, and its upper bound $b_{i}$ is set to the maximum of the
corresponding group upper bounds. Also, with this approach to ordinal
regression computation does not depend on the number of groups.

The predicted probability of an observation being in the $k$th group is given
by $\hat{p}(y\in k\,|\,\mathbf{x})=CDF(\hat{g}(b_{k}))-CDF(\hat{g}(a_{k}))$
where $a_{k}$ and $b_{k}$ are respectively the lower and upper bound specified
for the $k$th group on the chosen initial measurement scale. $CDF(z)$ is given
by (\ref{e20}) for symmetric models, and (\ref{e31}) for asymmetric models,
based on the OmniReg function estimates $\hat{g}(y)$, $\hat{f}(\mathbf{x})$,
and $\hat{s}(\mathbf{x})$ or $\hat{s}_{l}(\mathbf{x})$, $\hat{s}%
_{u}(\mathbf{x})$.

\section{Related work\label{rel}}

The OmniReg procedure presented here combines approaches from several separate
topics each of which has a long and rich history in Statistics, Econometrics
and Machine Learning. The procedure presented in Section \ref{est} is a
straight forward generalization of Tobit analysis (Tobin 1958) to incorporate
logistic errors, general censoring, heteroscedasticity, and general
nonparametric function estimation via boosted tree ensembles. There are many
previous works that incorporate various subsets of these generalizations.

\subsection{Censoring}

Although various forms of censoring actually occur in many types of applied
regression problems, it has been studied mainly in the context of survival
analysis where the outcome variable $y$\ is time to some event. Right
censoring occurs when that time exceeds the end of the study. The most popular
tool for survival analysis is the Cox proportional hazards model (Cox 1972)
which estimates the hazard function $\lambda(y)$ up to a multiplicative
function of time%
\[
\lambda(y\,|\,\mathbf{x})=p(y\,|\,\mathbf{x})/(1-CDF(y\,|\,\mathbf{x}%
))=h(y)\cdot\exp(f(\mathbf{x}))\text{.}%
\]
Here $h(y)$ is an unknown baseline hazard and the function $f(\mathbf{x})$ is
estimated from the data. Cox originally proposed a linear model $f(\mathbf{x}%
)=\mathbf{\beta}^{t}\mathbf{x}$, but various authors have since presented
nonlinear generalizations. OmniReg (Sections \ref{est} and \ref{opt}) can be
applied to survival data with mixtures of any type of censoring. Because it
estimates $p(y\,|\,\mathbf{x})$ it can be used to directly estimate the
absolute hazard $\lambda(y\,|\,\mathbf{x})$.

\subsection{Transformations}

Transforming an outcome variable $y$ in order to increase its compatibility
with regression procedure assumptions has a long history in Statistics.
Mosteller and Tukey (1977) proposed a ladder of reexpressions for different
variable types based upon their realizable values. Box and Cox (1964) proposed
the first data driven approach by selecting a power transformation $g_{\alpha
}(y)=y^{\alpha}$ for numeric variables that maximizes the data Gaussian
likelihood in the context of homoscedasticity and linear models.

Gifi (1990), and Breiman and Friedman (1985), proposed estimating outcome
transformations for additive modeling by minimizing%
\begin{equation}
\sum_{i=1}^{N}\left.  \left(  g(y_{i})-\sum_{j=1}^{p}h_{j}(x_{ij})\right)
^{2}\right/  \sum_{i=1}^{N}g^{2}(y_{i}) \label{e21}%
\end{equation}
jointly with respect to centered functions $g(y)$ and $\{h_{j}(x_{j}%
)\}_{1}^{p}$ under a smoothness constraint. With Gifi (1990) smoothness was
imposed by restricting all functions to be in the class of monotone splines
with a given number and placement of knots. Solving (\ref{e21}) then becomes a
canonical correlation problem implemented by alternating linear least--squares
parameter estimation. Breiman and Friedman (1985) directly used nonparametric
data smoothers to estimate all functions also using an alternating approach
similar to that used in Section \ref{opt}. \ Both approaches assume an
additive homoscedastic model for $g(y)$ as a function of $\mathbf{x}$.

Solutions to (\ref{e21}) maximize the correlation between $g(y)$ and
$\sum_{j=1}^{p}h_{j}(x_{j})$ over the joint distribution $p(\mathbf{x},y)$ of
$y$ and $\mathbf{x}$. As a result the solution transformations depend on the
marginal distribution $p(\mathbf{x})$ of the predictor variables $\mathbf{x}$
(Buja 1990). A consequence is that the expected solutions of (\ref{e21})
applied to data arising from the regression model
\[
g^{\ast}(y_{i})=\sum_{j=1}^{p}h_{j}^{\ast}(x_{ij})+\varepsilon_{i}%
\]
with $\varepsilon_{i}\sim N(0,\sigma^{2})$ will not generally be $g(y)=$
$g^{\ast}(y)$ and $\{h_{j}(x_{j})=h_{j}^{\ast}(x_{j})\}_{1}^{p}$.
Distributions of $\mathbf{x}$ that tend to exhibit clustering are especially problematic.

In order to alleviate this problem Tibshirani (1988) proposed finding
transformations $g(y)$ and $\{h_{j}(x_{j}\}_{1}^{p}$ that that minimize%
\[
\sum_{i=1}^{N}\left(  g(y_{i})-\sum_{j=1}^{p}h_{j}(x_{j})\right)  ^{2}%
\]
where the monotone function $g(y)$ is taken to be the variance stabilizing
transformation%
\begin{equation}
Var\left(  \left.  g(y)\,\right\vert \sum_{j=1}^{p}h_{j}(x_{j})\right)
=\text{ \emph{constant}.} \label{e22}%
\end{equation}
Solutions are obtained by an alternating optimization algorithm.

In the language of this paper (\ref{e22}) will hold if the location function
$f(\mathbf{x})$ (\ref{e10.5}) is additive in $\mathbf{x}$ and the scale
function $s(\mathbf{x})$ is independent of $f(\mathbf{x})$. In particular,
(\ref{e22}) does not necessarily imply homoscedasticity $s(\mathbf{x}%
)=\,$\emph{constant} in the transformed setting $g(y)$. For all the examples
in this paper the association between the location and scale estimates for the
optimal $\hat{g}(y)$ happened to be quite weak but there was still
considerable variation in scale, as with for example $\hat{s}_{l}(\mathbf{x})$
in Fig. \ref{fig32}.

The transformation strategy outlined in Sections \ref{est}, \ref{opt} and
\ref{asym} is directly regression based in that it is conditional on
$\mathbf{x}$. The basic assumption is that the standardized residuals
(\ref{e29}) (\ref{e35}) follow a standard logistic distribution (\ref{e2}) at
any $\mathbf{x}$ irrespective of the distribution of $\mathbf{x}$. Also note
that the estimate $\hat{g}(y)$ obtained from (\ref{e10.9}) is automatically monotonic.

\subsection{Quantile regression\label{quant}}

For known $y$-values and in the absence of censoring a natural alternative to
the procedures described in Sections \ref{est}, \ref{opt} and \ref{asym} is
quantile regression based on gradient boosted trees. One can estimate the
$p$th quantile function, $q_{p}(\mathbf{x)}$, of $p(y\,|\,\mathbf{x})$ using
the loss function
\begin{equation}
L(y,q_{p})=(y-q_{p})\,[p\,I(y\geq q_{p})+(p-1)\,I(y<q_{p})]\text{.}\label{e23}%
\end{equation}
Friedman (2001) suggested using (\ref{e23}) in the context of gradient
boosting and it has been incorporated in many implementations (see Ridgeway
2007 and Pedregosa 2011). Recently, Athey, Tibshirani, and Wagner (2017)
implemented quantile regression in random forests. With boosted tree quantile
regression (\ref{e23}) simply replaces (\ref{e3}) as the loss function in the
procedure outlined in Sections \ref{loss} and \ref{impl}, and only one
function is estimated separately for each quantile $p$. One can use this
procedure to independently estimate functions $q_{p}(\mathbf{x)}$ for several
quantiles $\{p_{k}\}_{1}^{P}$, and then \ for any $\mathbf{x}$ use
$\{q_{p}(\mathbf{x)\}}_{1}^{P}$ to approximate the corresponding quantiles of
$p(y\,|\,\mathbf{x})$. This approach makes no assumptions concerning a
generating model for $p(y\,|\,\mathbf{x})$, such as (\ref{e10.5}) or
(\ref{e30}). It thus may provide useful results in applications where those
assumptions are seriously violated. However, when they are not decreased
accuracy can result.

\subsection{Direct maximum likelihood estimation}

For $y$-values on a known measurement scale, another natural alternative to
the procedure described in Sections \ref{est}, \ref{opt} and \ref{asym} is
applying direct maximum likelihood such as in GAMLSS (Rigby and Stasinopoulos
2006). A parameterized probability density for $p(y\,|\,\mathbf{x})$ is
hypothesized and then its parameters are fitted as functions of the predictor
variables $\mathbf{x}$ by maximum likelihood. Parametric, linear and additive
functions are considered. These are fit to data by numerically maximizing the
corresponding (non convex) log--likelihood with respect to the parameters
defining all of the functions of $\mathbf{x}$. The assumed probability density
can have parameters controlling its location, scale, skewness and kurtosis
which are all modeled as functions of the predictors in this manner.

A similar paradigm is used in Sections \ref{est} and \ref{asym} based on the
two and three parameter logistic distributions. A difference is that the
parameters are taken to be more flexible functions of $\mathbf{x}$ as
represented by boosted decision tree ensembles. It is straightforward to
generalize the gradient tree boosting strategy of Section \ref{est} and
\ref{asym} to other distributions with more that two or three parameters as
functions of the predictor variables $\mathbf{x}$. Whether such added
complexity translates to improved or reduced accuracy will be application
dependent. In any application the diagnostics described in Sections
\ref{plots} and \ref{stdres} \ may be helpful in assessing model deficiencies.

The central difference between the GAMLSS method and OmniReg lies in the
optimal transformation strategy (Section \ref{opt}). The GAMLSS approach
assumes that the hypothesized parametric distribution describes
$p(y\,|\,\mathbf{x})$. OmniReg assumes that there exists some transformation
$g(y)$ for which $p(g(y)\,|\,\mathbf{x})$ follows the specified distribution.
It then nonparametrically estimates that transformation $g(y)$ jointly with
its corresponding distribution parameters as functions of $\mathbf{x}$. This
substantially enlarges the scope of application. The data examples of Sections
\ref{mash} and \ref{mds} illustrate the gains that \ can be made by this
transformation strategy. They also illustrate that the optimal transformations
for a given problem need not be close to any function in commonly assumed
parametric families such as the power family $g_{\alpha}(y)=y^{\alpha}$.

Arrays of multiple Q--Q plots each based on different regions of the
$\mathbf{x}$-space were proposed by van Buuren and Fredriks (2001). The
diagnostic plots of Section \ref{stdres} are a straightforward generalization.
Those of Section \ref{plots} can be viewed as a generalized version for
arbitrarily censored data.

\subsection{Ordinal regression}

There is a large literature on ordinal regression (see McCCullagh 1980 and
Guti\'{e}rrez \emph{et. al.} 2016). A common method is to apply a sequence of
binary linear logistic regressions to separately estimate the probability of
being below each of the successive group boundaries. The probability of being
within each of the groups is then obtained by successive differences between
these estimates. In order to avoid negative values the coefficients of all
linear models are constrained to be the same, with only the intercepts changing.

The approach most similar in spirit to the one here is Messner\emph{ et. al.}
(2014). They propose a statistical model similar to (\ref{e2}) (\ref{e10.5}).
The location $f(\mathbf{x})$ and log-scale functions $\log(s(\mathbf{x}))$ are
modeled as linear functions of the predictor variables $\mathbf{x}$ by
numerical maximum likelihood. The transformation $g(y)$ is preselected and not
systematically fit to the data.

\section{Point prediction\label{point}}

The focus of this work is on the estimation of the distribution functions
$p(y\,|\,\mathbf{x})$. These can be used to infer likely values of $y$ for
given $\mathbf{x}$-vectors. Sometimes the goal of regression is to return a
single value that minimizes the prediction risk at $\mathbf{x}$%
\begin{equation}
h(\mathbf{x})=\arg\min_{h}\int L(y,h)\,p(y\,|\,\mathbf{x})\,dy\text{.}%
\label{e40}%
\end{equation}
Here $L(y,h)$ represents a loss, relevant to the application at hand, for
predicting $h$ when the actual value is $y$. One way to attempt to solve
(\ref{e40}) is%
\begin{equation}
\hat{h}(\mathbf{x})=\arg\min_{h(\mathbf{x})\in H}\sum_{i=1}^{N}L(y_{i}%
,h(\mathbf{x}_{i}))\text{.}\label{e41}%
\end{equation}
Here $H$ is a class of functions usually defined by the fitting procedure
employed. An alternative strategy is to use%
\begin{equation}
\hat{h}(\mathbf{x})=\arg\min_{h}\int L(y,h)\,\hat{p}(y\,|\,\mathbf{x}%
)\,dy\label{e42}%
\end{equation}
where $\hat{p}(y\,|\,\mathbf{x})$ is an estimate of $p(y\,|\,\mathbf{x})$, for
example through (\ref{e2.5}) (\ref{e3}) (\ref{e4}). For some $L(y,h)$ and
$\hat{p}(y\,|\,\mathbf{x})$, (\ref{e42}) can be solved analytically. Otherwise
it can be easily solved by numerical methods. Note that there is no
requirement that $L(y,h)$ be convex in (\ref{e42}). Usually (\ref{e41}) is
used when there is no censoring, and (\ref{e42}) is applied when censoring is
present. Which of the two is best in terms of accuracy in any particular
application depends on the specific nature of that application. If there
exists a validation data set $V$ not used for training with known (uncensored)
$y$-values one can use%
\begin{equation}
CV=\sum_{i\in V}L(y_{i},\hat{h}(\mathbf{x}_{i}))\label{e43}%
\end{equation}
to estimate the accuracy of any $\hat{h}(\mathbf{x})$. An advantage of
(\ref{e42}) is that solutions for a variety of loss functions $L(y,h)$, as
well as other statistics, can be easily obtained without having to recompute
$\,\hat{p}(y\,|\,\mathbf{x})$.

\section{Discussion\label{disc}}

The goal of this work (OmniReg) is to provide a general unified procedure that
can be used for predictive inference in a variety of regression problems.
These include heteroscedasticity, asymmetry, ordinal regression, and general censoring.

Censoring has received relatively little attention in machine learning even
though it is a common occurrence in regression problems. Often there are
restrictions on the measured values of the outcome $y$. Section \ref{quest}
illustrates a case where measurements of a continuous outcome (age) are
restricted to six discrete values (intervals). Another common restriction is
where the observed $y$-values cannot be negative, such as payout on insurance
policy claims. There is usually a large mass of zero--valued outcomes along
with some positive ones. In this case it is not straight forward to formulate
and estimate a corresponding $p(y\,|\,\mathbf{x})$. One way (Tobin 1958) is to
consider the observed $y$-values as censored below zero measurements of a
latent variable $y^{\ast}$ with an unrestricted $p(y^{\ast}\,|\,\mathbf{x})$.
This can then be estimated from the data using the techniques of Sections
\ref{est}, \ref{opt} and \ref{asym}.

Formal predictive inference has also received relatively little attention.
However, informal inference is at the heart of prediction. The presumption is
that a predicted value $\hat{f}(\mathbf{x})$ is \textquotedblleft somewhere
close\textquotedblright\ to the actual outcome $y$-value. Predictive inference
simply quantifies this concept through an estimated probability distribution.
As seen in the data examples of Section \ref{examp} the nature of
\textquotedblleft somewhere close\textquotedblright\ can be very different for
different predictions in the same problem. Also this probability distribution
can be used to obtain point estimates for any loss function through
(\ref{e42}). These can sometimes be more accurate than corresponding direct
estimates (\ref{e41}).

A central feature of OmniReg is the optimal transformation strategy of Section
\ref{opt}. This extends the applicability of the method to a much wider class
of problems. It is also at the heart of the ordinal regression strategy of
Section \ref{ord}. As seen in the data examples, a particular assumed
probability distribution $p(y\,|\,\mathbf{x})$\ (here (\ref{e2.5}) or
(\ref{e30})) is seldom appropriate for the original outcome $y$. However,
there is often a corresponding optimal transformation $g(y)$ for which it is
much more appropriate. This optimal transformation may not be close to any
member of a common parametric family as was seen in Figs. \ref{fig12},
\ref{fig17} and \ref{fig30}.

The basic OmniReg method is indifferent to the technique used to obtain the
parameter function estimates $\hat{f}(\mathbf{x})$, $\hat{s}(\mathbf{x})$, or
$\hat{s}_{l}(\mathbf{x})$, $\hat{s}_{u}(\mathbf{x})$. An iterative strategy
based on gradient boosted trees was used in Sections \ref{impl}, \ref{opt} and
\ref{asym}. This seems to work well in a variety of applications. However,
there may be some for which other methods work as well or better. A crucial
but delicate ingredient is regularization, especially for the location
estimate. An overfitted location estimate $\hat{f}(\mathbf{x})$ will cause a
severe negative bias in the log--scale estimates $\log(\hat{s}(\mathbf{x})),$
or $\log(\hat{s}_{l}(\mathbf{x}))$, $\log(\hat{s}(\mathbf{x}))$. This produces
inaccurate (over optimistic) estimates of the corresponding scale functions,
which would be detected by the diagnostics of Section \ref{diag}. Early
stopping based on cross--validation seems to work well for the estimation
method used here. Other methods may require different regularization strategies.

Employing a different function estimation method changes the function class
$F$ in (\ref{e4}) or its asymmetric counterpart (Section \ref{asym}). This
criterion is minimized with respect to the transformation function $g(y)$
jointly with the parameter functions $f(\mathbf{x})$, $s(\mathbf{x})$, or
$s_{l}(\mathbf{x})$, $s_{u}(\mathbf{x})$ all in $F$. Thus, the expected
solution to (\ref{e10.5}) for $g(y)$ will depend on the function class $F$ as
specified by the function estimation procedure used. The optimal
transformation estimate $\hat{g}(y)$ will tend to be biased away from a
population optimal transformation towards those that permit more accurate
estimation of the corresponding parameters as functions of $\mathbf{x}$, using
the chosen regression method. In this sense the optimal transformation
strategy can somewhat compensate for less than perfect parameter function estimation.

The OmniReg method is also indifferent to the basic probability distribution
assumed for the error in the transformed setting. The logistic distribution
(\ref{e2}) \ is used here. It has high relative efficiency for narrow tailed
error distributions as well as being highly robust in the presence of
outliers. The strategy \ outlined in Sections \ref{est} -- \ref{asym} can
clearly be implemented using any assumed probability distribution. The optimal
transformation solutions (Section \ref{opt}) tend to be insensitive to
particular choices as long as their corresponding loss functions (negative
log--likelihood) are sufficiently robust. Inference in the transformed setting
$g(y)$ will however depend on a chosen distribution. In the absence of
censoring the standardized residual plots of Section \ref{stdres} can be used
to compare the actual transformed residuals to a spectrum of potential
candidate distributions as seen in the data examples of Sections \ref{mash}
and \ref{mds}. For censored data, predicted distributions for the marginal
distribution plots (Section \ref{plots}) can be constructed for any
distribution by substituting its cumulative distribution for (\ref{e20}).
These can then be compared to the actual empirical distribution of the
transformed data.

An essential ingredient of the OmniReg approach described here is the
diagnostic procedures presented in Section \ref{diag}. No predictive model
should ever be deployed without at least some assurance, beyond that of the
fitting procedure itself, that its predictions are valid. In the case
of\ point predictions, in the absence of censoring, (\ref{e1}) can be used as
a diagnostic to assess average error over all predictions based on $\hat
{f}(\mathbf{x})$. There is no such simple approach to assessing the accuracy
of individual probability distribution estimates $\hat{p}(y\,|\,\mathbf{x})$.
The diagnostic plots described in Section \ref{plots} for censored data, and
Section \ref{stdres} for uncensored data, represent one such approach. They
were able to expose the inadequacies of the untransformed solutions in all of
the data examples of Section \ref{examp}, and also detect shortcomings in the
symmetric transformed solution of Section \ref{mds}. Besides the procedures
described here, these diagnostics can be applied to other methods whose goal
is to estimate $p(y\,|\,\mathbf{x})$ and/or its associated parameters.

\section{Acknowledgement}

Enlightening discussions with Trevor Hastie are gratefully acknowledged.

\end{document}